# Active Tuples-based Scheme for Bounding Posterior Beliefs

**Bozhena Bidyuk**                                    BBIDYUK@GOOGLE.COM
*Google Inc.*
*19540 Jamboree Rd*
*Irvine, CA 92612*

**Rina Dechter**                                      DECHTER@ICS.UCI.EDU
*Donald Bren School of Information and Computer Science*
*University Of California Irvine*
*Irvine, CA 92697-3425*

**Emma Rollon**                                       EROLLON@LSI.UPC.EDU
*Departament de Llenguatges i Sistemes Informàtics*
*Universitat Politècnica de Catalunya*
*Barcelona, Spain*

## Abstract

The paper presents a scheme for computing lower and upper bounds on the posterior marginals in Bayesian networks with discrete variables. Its power lies in its ability to use any available scheme that bounds the probability of evidence or posterior marginals and enhance its performance in an anytime manner. The scheme uses the cutset conditioning principle to tighten existing bounding schemes and to facilitate anytime behavior, utilizing a fixed number of cutset tuples. The accuracy of the bounds improves as the number of used cutset tuples increases and so does the computation time. We demonstrate empirically the value of our scheme for bounding posterior marginals and probability of evidence using a variant of the bound propagation algorithm as a plug-in scheme.

## 1. Introduction

This paper addresses the problem of bounding the probability of evidence and posterior marginals in Bayesian networks with discrete variables. Deriving bounds on posteriors with a given accuracy is clearly an NP-hard problem (Abdelbar & Hedetniemi, 1998; Dagum & Luby, 1993; Roth, 1996) and indeed, most available approximation algorithms provide little or no guarantee on the quality of their approximations. Still, a few approaches were presented in the past few years for bounding posterior marginals (Horvitz, Suermondt, & Cooper, 1989; Poole, 1996, 1998; Mannino & Mookerjee, 2002; Mooij & Kappen, 2008) and for bounding the probability of evidence (Dechter & Rish, 2003; Larkin, 2003; Leisink & Kappen, 2003).

In this paper we develop a framework that can accept any bounding scheme and improve its bounds in an anytime manner using the cutset-conditioning principle (Pearl, 1988). To facilitate our scheme we develop an expression that converts bounds on the probability of evidence into bounds on posterior marginals.

Given a Bayesian network defined over a set of variables $\mathcal{X}$, a variable $X \in \mathcal{X}$, and a domain value $x \in \mathcal{D}(X)$, a posterior marginal $P(x|e)$ (where $e$ is a subset of assignments to the variables, called evidence) can be computed directly from two joint probabilities,

$P(x, e)$ and $P(e)$:

$$P(x|e) = \frac{P(x, e)}{P(e)} \qquad (1)$$

Given a set C=$\{C_1, ..., C_p\} \subset \mathcal{X}$ of cutset variables (e.g., a loop-cutset), we can compute the probability of evidence by enumerating over all the cutset tuples $c^i \in \mathcal{D}(C) = \prod_{i=1}^{p} D(C_i)$ using the formula:

$$P(e) = \sum_{i=1}^{M} P(c^i, e) \qquad (2)$$

where $M = |\mathcal{D}(C)|$. We can also compute the posterior marginals using the expression:

$$P(x|e) = \sum_{i=1}^{M} P(x|c^i, e) P(c^i|e) \qquad (3)$$

The computation of $P(c^i, e)$ for any assignment $c = c^i$ is linear in the network size if $C$ is a loop-cutset and it is exponential in $w$ if $C$ is a $w$-cutset (see definition in Section 2). The limitation of the cutset-conditioning method, as defined in Eq. (2) and (3), is that the number of cutset tuples $M$ grows exponentially with the cutset size.

There are two basic approaches for handling the combinatorial explosion in the cutset-conditioning scheme. One is to sample over the cutset space and subsequently approximate the distribution $P(C|e)$ from the samples, as shown by Bidyuk and Dechter (2007). The second approach, which we use here, is to enumerate $h$ out of $M$ tuples and bound the rest. We shall refer to the selected tuples as "active" tuples. A lower bound on $P(e)$ can be obtained by computing exactly the quantities $P(c^i, e)$ for $1 \le i \le h$ resulting in a partial sum in Eq. (2). This approach is likely to perform well if the selected $h$ tuples contain most of the probability mass of $P(e)$. However, this approach cannot be applied directly to obtain the bounds on the posterior marginals in Eq. (3). Even a partial sum in Eq. (3) requires computing $P(c^i|e)$ which in turn requires a normalization constant $P(e)$. We can obtain naive bounds on posterior marginals from Eq. (1) using $P^L(e)$ and $P^U(e)$ to denote available lower and upper bounds over joint probabilities:

$$\frac{P^L(x, e)}{P^U(e)} \le P(x|e) \le \frac{P^U(x, e)}{P^L(e)}$$

However, those bounds usually perform very poorly and often yield an upper bound $> 1$.

Horvitz et. al (1989) were the first to propose a scheme for bounding posterior marginals based on a subset of cutset tuples. They proposed to select $h$ highest probability tuples from $P(c)$ and derived lower and upper bounds on the sum in Eq. (3) from the joint probabilities $P(c^i, e)$ and priors $P(c^i)$ for $1 \le i \le h$. Their resulting *bounded conditioning* algorithm was shown to compute good bounds on the posterior marginals of some variables in an Alarm network (with $M = 108$). However, the intervals between lower and upper bound values increase as the probability of evidence becomes smaller because the prior distribution becomes a bad predictor of the high probability tuples in $P(C|e)$ and $P(c)$ becomes a bad upper bound for $P(c, e)$.

The expression we derive in this paper yields a significantly improved formulation which results in our *Active Tuples Bounds (ATB) framework*. The generated bounds facilitate anytime performance and are provably tighter than the bounds computed by bounded conditioning. In addition, our expression accommodates the use of any off-the-shelf scheme which bounds the probability of evidence. Namely, $ATB$ accepts any algorithm for bounding $P(e)$ and generates an algorithm that bounds the posterior marginals. Moreover, it can also tighten the input bounds on $P(e)$.

The time complexity of $ATB$ is linear in the number of active (explored) cutset tuples $h$. If the complexity of bounding $P(e)$ is $O(T)$, then bounding the probability mass of the unexplored tuples is $O(T \cdot h \cdot (d-1) \cdot |C|)$ where $|C|$ is the number of variables in the cutset and $d$ is the maximum domain size.

We evaluate our framework experimentally, using a variant of *bound propagation* ($BdP$) (Leisink & Kappen, 2003) as the plug-in bounding scheme. $BdP$ computes bounds by iteratively solving a linear optimization problem for each variable where the minimum and maximum of the objective function correspond to lower and upper bounds on the posterior marginals. The performance of $BdP$ was demonstrated on the Alarm network, the Ising grid network, and on regular bipartite graphs. Since bound propagation is exponential in the Markov boundary size, and since it requires solving linear programming problems many times, its overhead as a plug-in scheme was too high and not cost-effective. We therefore utilize a variant of bound propagation called $ABdP+$, introduced by Bidyuk and Dechter (2006b), that trades accuracy for speed.

We use Gibbs cutset sampling (Bidyuk & Dechter, 2003a, 2003b) for finding high-probability cutset tuples. Other schemes, such as stochastic local search (Kask & Dechter, 1999) can also be used. The investigation into generating high-probability cutset tuples is outside the primary scope of the paper.

We show empirically that $ATB$ using bound propagation is often superior to bound propagation alone when both are given comparable time resources. More importantly, $ATB$'s accuracy improves with time. We also demonstrate the power of $ATB$ for improving the bounds on probability of evidence. While the latter is not the main focus of our paper, lower and upper bounds on the probability of evidence are contained in the expression for bounding posterior marginals.

The paper is organized as follows. Section 2 provides background on the previously proposed method of bounded conditioning. Section 3 presents and analyzes our $ATB$ framework. Section 4 describes the implementation details of using bound propagation as an $ATB$ plug-in and presents our empirical evaluation. Section 5 discusses related work, and Section 6 concludes.

## 2. Background

For background, we define key concepts and describe the bounded conditioning algorithm which inspired our work.

### 2.1 Preliminaries

In this section, we define essential terminology and provide background information on Bayesian networks.

DEFINITION **2.1 (graph concepts)** *A **directed graph** is a pair $\mathcal{G} = <\mathcal{V}, \mathcal{E}>$, where $\mathcal{V} = \{X_1, ..., X_n\}$ is a set of nodes and $\mathcal{E} = \{(X_i, X_j) | X_i, X_j \in \mathcal{V}\}$ is the set of edges. Given $(X_i, X_j) \in \mathcal{E}$, $X_i$ is called a **parent** of $X_j$, and $X_j$ is called a **child** of $X_i$. The set of $X_i$'s parents is denoted $pa(X_i)$, or $pa_i$, while the set of $X_i$'s children is denoted $ch(X_i)$, or $ch_i$. The family of $X_i$ includes $X_i$ and its parents. The **moral graph** of a directed graph $\mathcal{G}$ is the undirected graph obtained by connecting the parents of each of the nodes in $\mathcal{G}$ and removing the arrows. A **cycle-cutset** of an undirected graph is a subset of nodes that, when removed, yields a graph without cycles. A **loop** in a directed graph $\mathcal{G}$ is a subgraph of $\mathcal{G}$ whose underlying graph is a cycle (undirected). A directed graph is acyclic if it has no directed loops. A directed graph is **singly-connected** (also called a **poly-tree**), if its underlying undirected graph has no cycles. Otherwise, it is called **multiply-connected**.*

DEFINITION **2.2 (loop-cutset)** *A vertex $v$ is a **sink** with respect to a loop $\mathcal{L}$ if the two edges adjacent to $v$ in $\mathcal{L}$ are directed into $v$. A vertex that is not a sink with respect to a loop $\mathcal{L}$ is called an **allowed** vertex with respect to $\mathcal{L}$. A **loop-cutset** of a directed graph $\mathcal{G}$ is a set of vertices that contains at least one allowed vertex with respect to each loop in $\mathcal{G}$.*

DEFINITION **2.3 (Bayesian network)** *Let $\mathcal{X} = \{X_1, ..., X_n\}$ be a set of random variables over multi-valued domains $\mathcal{D}(X_1), ..., \mathcal{D}(X_n)$. A **Bayesian network** $\mathcal{B}$ (Pearl, 1988) is a pair $<\mathcal{G}, \mathcal{P}>$ where $\mathcal{G}$ is a directed acyclic graph whose nodes are the variables $\mathcal{X}$ and $\mathcal{P} = \{P(X_i | pa_i) \mid i = 1, ..., n\}$ is the set of conditional probability tables (CPTs) associated with each $X_i$. $\mathcal{B}$ represents a joint probability distribution having the product form:*

$$P(x_1, ...., x_n) = \prod_{i=1}^{n} P(x_i | pa(X_i))$$

*An evidence $e$ is an instantiated subset of variables $E \subset \mathcal{X}$.*

DEFINITION **2.4 (Markov blanket and Markov boundary)** *A **Markov blanket** of $X_i$ is a subset of variables $Y \subset \mathcal{X}$ such that $X_i$ is conditionally independent of all other variables given $Y$. A **Markov boundary** of $X_i$ is its minimal Markov blanket (Pearl, 1988).*

In our following discussion we will identify Markov boundary $X_i$ with the Markov blanket consisting of $X_i$'s parents, children, and parents of its children.

DEFINITION **2.5 (Relevant Subnetwork)** *Given evidence $e$, **relevant subnetwork** of $X_i$ relativde to $e$ is a subnetwork of $\mathcal{B}$ obtained by removing all descendants of $X_i$ that are not observed and do not have observed descendants.*

If the observations change, the Markov boundary of $X_i$ will stay the same while its relevant subnetwork may change. As most inference tasks are defined relative to a specific set of observations $e$, it is often convenient to restrict attention to the Markov boundary of $X_i$ in the relevant subnetwork of $X_i$.

The most common query over Bayesian networks is *belief updating* which is the task of computing the posterior distribution $P(X_i | e)$ given evidence $e$ and a query variable $X_i \in \mathcal{X}$. Another query is to compute probability of evidence $P(e)$. Both tasks are NP-hard (Cooper, 1990). Finding approximate posterior marginals with a fixed accuracy is also NP-hard (Dagum & Luby, 1993; Abdelbar & Hedetniemi, 1998). When the network is a poly-tree, belief updating and other inference tasks can be accomplished in time linear in the size of the network. In general, exact inference is exponential in the induced width of the network's moral graph.

**DEFINITION 2.6 (induced width)** *The* width *of a node in an ordered undirected graph is the number of the node's neighbors that precede it in the ordering. The* width *of an ordering o, denoted $w(o)$, is the width over all nodes. The* induced width *of an ordered graph, $w^*(o)$, is the width of the ordered graph obtained by processing the nodes from last to first.*

**DEFINITION 2.7 ($w$-cutset)** *A $w$-cutset of a Bayesian network $\mathcal{B}$ is a subset of variables $C$ such that, when removed from the moral graph of the network, its induced width is $\leq w$.*

Throughout the paper, we will consider a Bayesian network over a set of variables $\mathcal{X}$, evidence variables $E \subset \mathcal{X}$ and evidence $E = e$, and a cutset $C = \{C_1, ..., C_p\} \subset \mathcal{X} \backslash E$. Lower-case $c = \{c_1, ..., c_p\}$ will denote an arbitrary instantiation of cutset $C$, and $M = |\mathcal{D}(C)| = \prod_{C_i \in C} |\mathcal{D}(C_i)|$ will denote the number of different cutset tuples.

## 2.2 Bounded Conditioning

Bounded conditioning ($BC$) is an anytime scheme for computing posterior bounds in Bayesian networks proposed by Horvitz et. al (1989). It is derived from the loop-cutset conditioning method (see Eq. 3). Given a node $X \in \mathcal{X}$ and a domain value $x \in \mathcal{D}(X)$, they derive the following bounds from the following formula:

$$P(x|e) = \sum_{i=1}^{M} P(x|c^i, e)P(c^i|e) = \sum_{i=1}^{h} P(x|c^i, e)P(c^i|e) + \sum_{i=h+1}^{M} P(x|c^i, e)P(c^i|e) \quad (4)$$

The hard-to-compute $P(c^i|e)$ is replaced for $i \leq h$ with a normalization formula:

$$P(x|e) = \frac{\sum_{i=1}^{h} P(x|c^i, e)P(c^i, e)}{\sum_{i=1}^{h} P(c^i, e) + \sum_{i=h+1}^{M} P(c^i, e)} + \sum_{i=h+1}^{M} P(x|c^i, e)P(c^i|e) \quad (5)$$

$BC$ computes exactly $P(c^i, e)$ and $P(x|c^i, e)$ for the $h$ cutset tuples and bounds the rest.

The lower bound is obtained from Eq. (5) by replacing $\sum_{i=h+1}^{M} P(c^i, e)$ in the denominator with the sum of priors $\sum_{i=h+1}^{M} P(c^i)$ and simply dropping the sum on the right:

$$P_{BC}^L(x|e) \triangleq \frac{\sum_{i=1}^{h} P(x, c^i, e)}{\sum_{i=1}^{h} P(c^i, e) + \sum_{i=h+1}^{M} P(c^i)} \quad (6)$$

The upper bound is obtained from Eq. (5) by replacing $\sum_{i=h+1}^{M} P(c^i, e)$ in the denominator with a zero, and replacing $P(x|c^i, e)$ and $P(c^i|e)$ for $i > h$ with the upper bounds of 1 and a derived upper bound (not provided here) respectively:

$$P_{BC}^U(x|e) \triangleq \frac{\sum_{i=1}^{h} P(x, c^i, e)}{\sum_{i=1}^{h} P(c^i, e)} + \frac{\sum_{i=h+1}^{M} P(c^i)}{\sum_{i=1}^{h} P^L(c^i|e) + 1 - \sum_{i=1}^{h} P^U(c^i|e)}$$

Applying definitions for $P^L(c^i|e) = \frac{P(c^i,e)}{\sum_{i=1}^h P(c^i,e) + \sum_{i=h+1}^M P(c^i)}$ and $P^U(c^i|e) = \frac{P(c^i,e)}{\sum_{i=1}^h P(c^i,e)}$ from Horvitz et al. (1989), we get:

$$P^U_{BC}(x|e) \triangleq \frac{\sum_{i=1}^h P(x,c^i,e)}{\sum_{i=1}^h P(c^i,e)} + \frac{(\sum_{i=h+1}^M P(c^i))(\sum_{i=1}^h P(c^i,e) + \sum_{i=h+1}^M P(c^i))}{\sum_{i=1}^h P(c^i,e)} \quad (7)$$

The bounds expressed in Eq. (6) and (7) converge to the exact posterior marginals as $h \to M$. However, we can show that,

THEOREM **2.1 (bounded conditioning bounds interval)** *The interval between lower and upper bounds computed by bounded conditioning is lower bounded by the probability mass of prior distribution $P(C)$ of the unexplored cutset tuples:*

$$\forall h, P^U_{BC}(x|e) - P^L_{BC}(x|e) \geq \sum_{i=h+1}^M P(c^i)$$

PROOF. See Appendix A. □

## 3. Architecture for Active Tuples Bounds

In this section, we describe our Active Tuples Bounds ($ATB$) framework. It builds on the same principles as bounded conditioning. Namely, given a cutset $C$ and some method for generating $h$ cutset tuples, the probabilities $P(c,e)$ of the $h$ tuples are evaluated exactly and the rest are upper and lower bounded. The worst bounds on $P(c,e)$ are the lower bound of 0 and the upper bound of $P(c)$. $ATB$ bounds can be improved by using a plug-in algorithm that computes tighter bounds on the participating joint probabilities. It always computes tighter bounds than bounded conditioning, even when using 0 and $P(c)$ to bound $P(c,e)$.

For the rest of the section, $c_{1:q} = \{c_1, ..., c_q\}$ with $q < |C|$ denotes a generic partial instantiation of the first $q$ variables in $C$, while $c^i_{1:q}$ indicates a particular partial assignment.

Given $h$ cutset tuples, $0 \leq h \leq M$, that we assume without loss of generality to be the first $h$ tuples according to some enumeration order, a variable $X \in \mathcal{X}\backslash E$ and $x \in \mathcal{D}(X)$, we can rewrite Eq. (3) as:

$$P(x|e) = \frac{\sum_{i=1}^M P(x,c^i,e)}{\sum_{i=1}^M P(c^i,e)} = \frac{\sum_{i=1}^h P(x,c^i,e) + \sum_{i=h+1}^M P(x,c^i,e)}{\sum_{i=1}^h P(c^i,e) + \sum_{i=h+1}^M P(c^i,e)} \quad (8)$$

The probabilities $P(x,c^i,e)$ and $P(c^i,e)$, $1 \leq i \leq h$, can be computed in polynomial time if $C$ is a loop-cutset and in time and space exponential in $w$ if $C$ is a $w$-cutset. The question is how to compute or bound $\sum_{i=h+1}^M P(x,c^i,e)$ and $\sum_{i=h+1}^M P(c^i,e)$ in an efficient manner.

Our approach first replaces the sums over the tuples $c^{h+1},...,c^M$ with a sum over a polynomial number (in $h$) of partially-instantiated tuples. From that, we develop new expressions for lower and upper bounds on the posterior marginals as a function of the lower and upper bounds on the joint probabilities $P(x, c_{1:q}, e)$ and $P(c_{1:q}, e)$. We assume in our derivation that there is an algorithm $\mathcal{A}$ that can compute those bounds, and refer to them as $P^L_{\mathcal{A}}(x, c_{1:q}, e)$ (resp. $P^L_{\mathcal{A}}(c_{1:q}, e)$) and $P^U_{\mathcal{A}}(x, c_{1:q}, e)$ (resp. $P^U_{\mathcal{A}}(c_{1:q}, e)$) respectively.

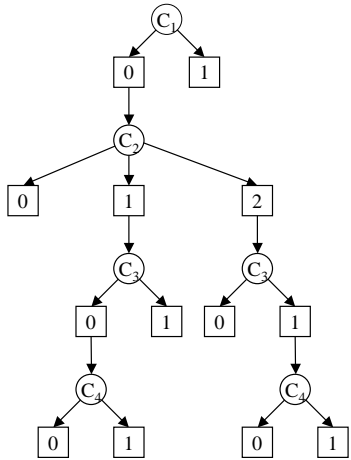

Figure 1: A search tree for cutset $C = \{C_1, ..., C_4\}$.

## 3.1 Bounding the Number of Processed Tuples

We will now formally define partially-insantiated tuples and replace the sum over the exponential number of uninstantiated tuples ($h + 1$ through $M$) with a sum over polynomial number of partially-instantiated tuples ($h + 1$ through $M'$) in Eq. 8.

Consider a fully-expanded search tree of depth $|C|$ over the cutset search space expanded in the order $C_1, ..., C_p$. A path from the root to the leaf at depth $|C|$ corresponds to a full cutset tuple. We will call such a path an *active path* and the corresponding tuple an *active tuple*. We can obtain the *truncated search tree* by trimming all branches that are not on the active paths:

DEFINITION **3.1 (truncated search tree)** *Given a search tree $T$ covering the search space $\mathcal{H}$ over variables $\mathcal{Y} = \{Y_1, ..., Y_m\} \subseteq \mathcal{X}$, a **truncated search tree** relative to a subset $S = \{y^1, ..., y^t\} \subset \mathcal{D}(Y_1) \times ... \times \mathcal{D}(Y_m)$ of full assignments, is obtained by marking the edges on all the paths appearing in $S$ and removing all unmarked edges and nodes except those emanating from marked nodes.*

Let $S = \{c^1, ..., c^h\}$. Clearly, the leaves at depth $q < |C|$ in the truncated search tree relative to $S$ correspond to the partially instantiated cutset tuples $c_{1:q}$ which are not extended to full cutset assignments.

EXAMPLE **3.1** *Consider a Bayesian network $\mathcal{B}$ with cutset variables $C = \{C_1, ..., C_4\}$, domain values $\mathcal{D}(C_1) = \mathcal{D}(C_3) = \mathcal{D}(C_4) = \{0, 1\}$, $\mathcal{D}(C_2) = \{0, 1, 2\}$, and four fully-instantiated tuples $\{0, 1, 0, 0\}$, $\{0, 1, 0, 1\}$, $\{0, 2, 1, 0\}$, $\{0, 2, 1, 1\}$. Figure 1 shows its truncated search tree, where the remaining partially instantiated tuples are $\{0, 0\}$, $\{0, 1, 1\}$, $\{0, 2, 0\}$, and $\{1\}$.*

PROPOSITION **3.1** *Let $C$ be a cutset, $d$ be the maximum domain size, and $h$ be the number of generated cutset tuples. Then the number of partially-instantiated cutset tuples in the truncated search tree is bounded by $O(h \cdot (d - 1) \cdot |C|)$.*

PROOF. Since every node in the path from the root $C_1$ to a leaf $C_p$ can not have more than $(d-1)$ emanating leaves, the theorem clearly holds. □

Let $M'$ be the number of truncated tuples. We can enumerate the partially instantiated tuples, denoting the $j$-th tuple as $c_{1:q_j}^j$, $1 \leq j \leq M'$, where $q_j$ is the tuple's length. Clearly, the probability mass over the cutset tuples $c^{h+1}, ..., c^M$ can be captured by a sum over the truncated tuples. Namely:

PROPOSITION **3.2**

$$\sum_{i=h+1}^{M} P(c^i, e) = \sum_{j=1}^{M'} P(c_{1:q_j}^j, e) \tag{9}$$

$$\sum_{i=h+1}^{M} P(x, c^i, e) = \sum_{j=1}^{M'} P(x, c_{1:q_j}^j, e) \tag{10}$$

□

Therefore, we can bound the sums over the tuples $h+1$ through $M$ in Eq. (8) by bounding a polynomial (in $h$) number of partially-instantiated tuples as follows,

$$P(x|e) = \frac{\sum_{i=1}^{h} P(x, c^i, e) + \sum_{j=1}^{M'} P(x, c_{1:q_j}^j, e)}{\sum_{i=1}^{h} P(c^i, e) + \sum_{j=1}^{M'} P(c_{1:q_j}^j, e)} \tag{11}$$

## 3.2 Bounding the Probability over the Truncated Tuples

In the following, we develop lower and upper bound expressions used by $ATB$.

### 3.2.1 LOWER BOUNDS

First, we decompose $P(c_{1:q_j}^j, e)$, $0 \leq j \leq M'$, as follows. Given a variable $X \in \mathcal{X}$ and a distinguished value $x \in \mathcal{D}(X)$:

$$P(c_{1:q_j}^j, e) = \sum_{x' \in D(X)} P(x', c_{1:q_j}^j, e) = P(x, c_{1:q_j}^j, e) + \sum_{x' \neq x} P(x', c_{1:q_j}^j, e) \tag{12}$$

Replacing $P(c_{1:q_j}^j, e)$ in Eq. (11) with the right-hand side of Eq. (12), we get:

$$P(x|e) = \frac{\sum_{i=1}^{h} P(x, c^i, e) + \sum_{j=1}^{M'} P(x, c_{1:q_j}^j, e)}{\sum_{i=1}^{h} P(c^i, e) + \sum_{j=1}^{M'} P(x, c_{1:q_j}^j, e) + \sum_{j=1}^{M'} \sum_{x' \neq x} P(x', c_{1:q_j}^j, e)} \tag{13}$$

We will use the following two lemmas:

LEMMA **3.1** *Given positive numbers $a > 0$, $b > 0$, $\delta \geq 0$, if $a < b$, then: $\frac{a}{b} \leq \frac{a+\delta}{b+\delta}$.* □

Lemma **3.2** *Given positive numbers $a$, $b$, $\delta$, $\delta^L$, $\delta^U$, if $a < b$ and $\delta^L \leq \delta \leq \delta^U$, then:*

$$\frac{a + \delta^L}{b + \delta^L} \leq \frac{a + \delta}{b + \delta} \leq \frac{a + \delta^U}{b + \delta^U}$$

$\square$

The proof of both lemmas is straight forward.

Lemma 3.2 says that if the sums in the numerator and denominator have some component $\delta$ in common, then replacing $\delta$ with a larger value $\delta^U$ in both the numerator and the denominator yields a larger fraction. Replacing $\delta$ with a smaller value $\delta^L$ in both places yields a smaller fraction.

Observe now that in Eq. (13) the sums in both the numerator and the denominator contain $P(x, c^j_{1:q_j}, e)$. Hence, we can apply Lemma 3.2. We will obtain a lower bound by replacing $P(x, c^j_{1:q_j}, e)$, $1 \leq j \leq M'$, in Eq. (13) with corresponding lower bounds in both numerator and denominator, yielding:

$$P(x|e) \geq \frac{\sum_{i=1}^{h} P(x, c^i, e) + \sum_{j=1}^{M'} P^L_{\mathcal{A}}(x, c^j_{1:q_j}, e)}{\sum_{i=1}^{h} P(c^i, e) + \sum_{j=1}^{M'} P^L_{\mathcal{A}}(x, c^j_{1:q_j}, e) + \sum_{j=1}^{M'} \sum_{x' \neq x} P(x', c^j_{1:q_j}, e)} \tag{14}$$

Subsequently, grouping $P^L_{\mathcal{A}}(x, c^j_{1:q_j}, e)$ and $\sum_{x' \neq x} P(x', c^j_{1:q_j}, e)$ under one sum and replacing $P^L_{\mathcal{A}}(x, c^j_{1:q_j}, e) + \sum_{x' \neq x} P(x', c^j_{1:q_j}, e)$ with its corresponding upper bound (increasing denominator), we obtain:

$$P(x|e) \geq \frac{\sum_{i=1}^{h} P(x, c^i, e) + \sum_{j=1}^{M'} P^L_{\mathcal{A}}(x, c^j_{1:q_j}, e)}{\sum_{i=1}^{h} P(c^i, e) + \sum_{j=1}^{M'} UB[P^L_{\mathcal{A}}(x, c^j_{1:q_j}, e) + \sum_{x' \neq x} P(x', c^j_{1:q_j}, e)]} \triangleq P^L_{\mathcal{A}}(x|e) \tag{15}$$

where upper bound UB can be obtained as follows:

$$UB[P^L_{\mathcal{A}}(x, c^j_{1:q_j}, e) + \sum_{x' \neq x} P(x', c^j_{1:q_j}, e)] \triangleq \min \begin{cases} P^L_{\mathcal{A}}(x, c^j_{1:q_j}, e) + \sum_{x' \neq x} P^U_{\mathcal{A}}(x', c^j_{1:q_j}, e) \\ P^U_{\mathcal{A}}(c^j_{1:q_j}, e) \end{cases} \tag{16}$$

The value $\sum_{x' \neq x} P^U_{\mathcal{A}}(x', c^j_{1:q_j}, e)$ is, obviously, an upper bound of $\sum_{x' \neq x} P(x', c^j_{1:q_j}, e)$. The value $P^U_{\mathcal{A}}(c^j_{1:q_j}, e)$ is also an upper bound since $P^L_{\mathcal{A}}(x, c^j_{1:q_j}, e) + \sum_{x' \neq x} P(x', c^j_{1:q_j}, e) \leq P(c^j_{1:q_j}, e) \leq P^U_{\mathcal{A}}(c^j_{1:q_j}, e)$. Neither bound expression in Eq. (16) dominates the other. Thus, we compute the minimum of the two values.

Please note that the numerator in Eq. (15) above also provides an anytime lower bound on the joint probability $P(x, e)$ and can be used to compute a lower bound on the probability of evidence. In general, a lower bound denoted $P_{\mathcal{A}}^{L}(e)$ is obtained by:

$$P(e) \geq \sum_{i=1}^{h} P(c^i, e) + \sum_{j=1}^{M'} P_{\mathcal{A}}^{L}(c_{1:q_j}^{j}, e) \triangleq P_{\mathcal{A}}^{L}(e) \tag{17}$$

### 3.2.2 Upper Bound

The upper bound expression can be obtained in a similar manner. Since both numerator and denominator in Eq. (13) contain addends $P(x, c_{1:q_j}^{j}, e)$, using Lemma 3.2 we replace each $P(x, c_{1:q_j}^{j}, e)$ with an upper bound $P_{\mathcal{A}}^{U}(x, c_{1:q_j}^{j}, e)$ yielding:

$$P(x|e) \leq \frac{\displaystyle\sum_{i=1}^{h} P(x, c^i, e) + \sum_{j=1}^{M'} P_{\mathcal{A}}^{U}(x, c_{1:q_j}^{j}, e)}{\displaystyle\sum_{i=1}^{h} P(c^i, e) + \sum_{j=1}^{M'} P_{\mathcal{A}}^{U}(x, c_{1:q_j}^{j}, e) + \sum_{j=1}^{M'} \sum_{x' \neq x} P(x', c_{1:q_j}^{j}, e)} \tag{18}$$

Subsequently, replacing each $P(x', c_{1:q_j}^{j}, e)$, $x' \neq x$, with a lower bound (reducing denominator), we obtain a new upper bound expression on $P(x|e)$:

$$P(x|e) \leq \frac{\displaystyle\sum_{i=1}^{h} P(x, c^i, e) + \sum_{j=1}^{M'} P_{\mathcal{A}}^{U}(x, c_{1:q_j}^{j}, e)}{\displaystyle\sum_{i=1}^{h} P(c^i, e) + \sum_{j=1}^{M'} P_{\mathcal{A}}^{U}(x, c_{1:q_j}^{j}, e) + \sum_{j=1}^{M'} \sum_{x' \neq x} P_{\mathcal{A}}^{L}(x', c_{1:q_j}^{j}, e)} \triangleq P_{\mathcal{A}}^{U}(x|e) \tag{19}$$

Similar to the lower bound, the numerator in the upper bound expression $P_{\mathcal{A}}^{U}(x|e)$ provides an anytime upper bound on the joint probability $P(x, c^i, e)$ which can be generalized to upper bound the probability of evidence:

$$P(e) \leq \sum_{i=1}^{h} P(c^i, e) + \sum_{j=1}^{M'} P_{\mathcal{A}}^{U}(c_{1:q_j}^{j}, e) \triangleq P_{\mathcal{A}}^{U}(e) \tag{20}$$

The derived bounds $P_{\mathcal{A}}^{L}(x|e)$ and $P_{\mathcal{A}}^{U}(x|e)$ are never worse than those obtained by bounded conditioning, as we will show in Section 3.4.

### 3.3 Algorithmic Description

Figure 2 summarizes the active tuples-based bounding scheme $ATB$. In steps 1 and 2, we generate $h$ fully-instantiated cutset tuples and compute exactly the probabilities $P(c^i, e)$ and $P(X, c^i, e)$ for $i \leq h$, $\forall X \in \mathcal{X} \backslash (C \cup E)$, using, for example, the *bucket-elimination* algorithm (Dechter, 1999). In step 3, we compute bounds on the partially instantiated tuples using algorithm $\mathcal{A}$. In step 4, we compute the lower and upper bounds on the posterior marginals using expressions (15) and (19), respectively.

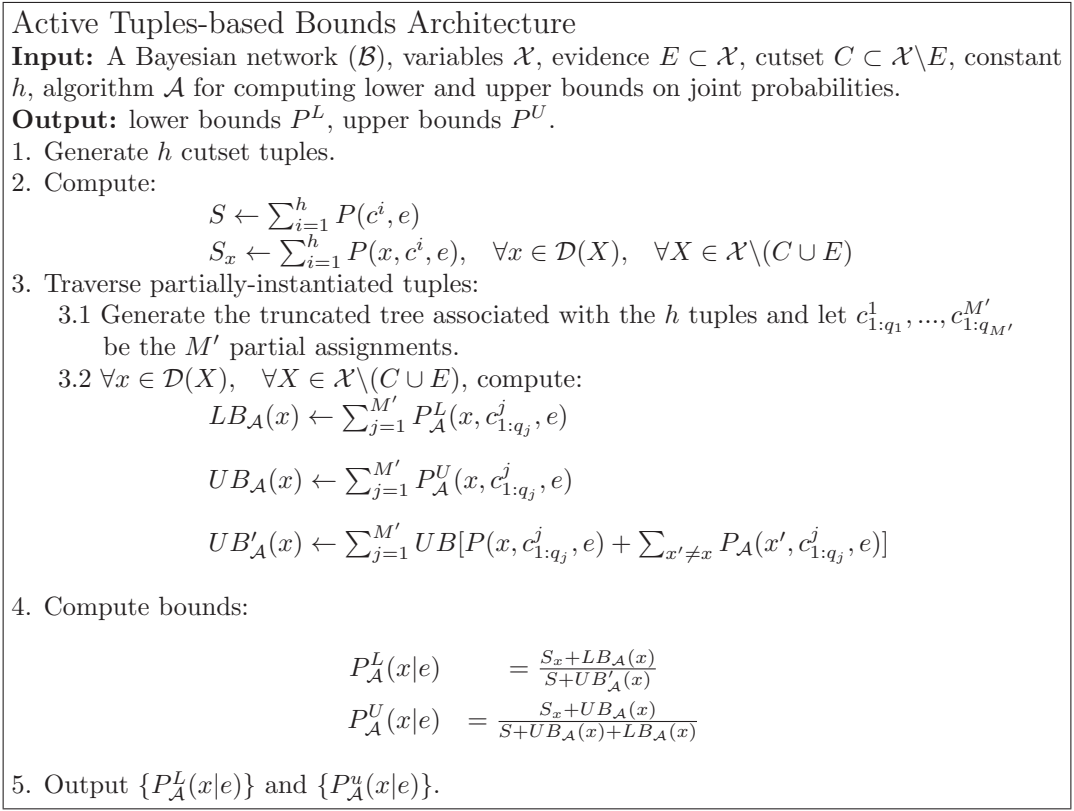

Figure 2: Active Tuples Bounds Architecture

EXAMPLE **3.2** *Consider again the Bayesian network $\mathcal{B}$ described in Example 3.1. Recall that $\mathcal{B}$ has a cutset $C = \{C_1, ..., C_4\}$ with domains $\mathcal{D}(C_1) = \mathcal{D}(C_3) = \mathcal{D}(C_4) = \{0, 1\}$ and $\mathcal{D}(C_2) = \{0, 1, 2\}$. The total number of cutset tuples is $M = 24$. Let $X \notin C$ be a variable in $\mathcal{B}$ with domain $\mathcal{D}(X) = \{x, x'\}$. We will compute bounds on $P(x|e)$. Assume we generated the same four cutset tuples ($h = 4$) as before:*

$$c^1 = \{C_1 = 0, C_2 = 1, C_3 = 0, C_4 = 0\} = \{0, 1, 0, 0\}$$
$$c^2 = \{C_1 = 0, C_2 = 1, C_3 = 0, C_4 = 1\} = \{0, 1, 0, 1\}$$
$$c^3 = \{C_1 = 0, C_2 = 2, C_3 = 1, C_4 = 0\} = \{0, 2, 1, 0\}$$
$$c^4 = \{C_1 = 0, C_2 = 2, C_3 = 1, C_4 = 1\} = \{0, 2, 1, 1\}$$

*The corresponding truncated search tree is shown in Figure 1. For the tuple $\{0, 1, 0, 0\}$, we compute exactly the probabilities $P(x, C_1 = 0, C_2 = 1, C_3 = 0, C_4 = 0, e)$ and $P(C_1 = 0, C_2 = 1, C_3 = 0, C_4 = 0)$. Similarly, we obtain exact probabilities $P(x, C_1 = 0, C_2 = 1, C_3 = 0, C_4 = 1)$ and $P(C_1 = 0, C_2 = 1, C_3 = 0, C_4 = 1)$ for the second cutset instance $\{0, 1, 0, 1\}$.*

Since $h = 4$, $\sum_{i=1}^{h} P(x', c^i, e)$ and $\sum_{i=1}^{h} P(c^i, e)$ are:

$$\sum_{i=1}^{4} P(x, c^i, e) = P(x, c^1, e) + P(x, c^2, e) + P(x, c^3, e) + P(x, c^4, e)$$

$$\sum_{i=1}^{4} P(c^i, e) = P(c^1, e) + P(c^2, e) + P(c^3, e) + P(c^4, e)$$

The remaining partial tuples are: $c_{1:2}^1 = \{0, 0\}$, $c_{1:3}^2 = \{0, 1, 1\}$, $c_{1:3}^3 = \{0, 2, 0\}$, and $c_{1:1}^4 = \{1\}$. Since these 4 tuples are not full cutsets, we compute bounds on their joint probabilities. Using the same notation as in Figure 2, the sums over the partially instantiated tuples will have the form:

$$UB_{\mathcal{A}}(x) \triangleq P_{\mathcal{A}}^U(x, c_{1:2}^1, e) + P_{\mathcal{A}}^U(x, c_{1:3}^2, e) + P_{\mathcal{A}}^U(x, c_{1:3}^3, e) + P_{\mathcal{A}}^U(x, c_{1:1}^4, e)$$

$$LB_{\mathcal{A}}(x) \triangleq P_{\mathcal{A}}^L(x, c_{1:2}^1, e) + P_{\mathcal{A}}^L(x, c_{1:3}^2, e) + P_{\mathcal{A}}^L(x, c_{1:3}^3, e) + P_{\mathcal{A}}^L(x, c_{1:1}^4, e)$$

From Eq. (19) we get:

$$P_{\mathcal{A}}^U(x|e) = \frac{\sum_{i=1}^{4} P(x, c^i, e) + UB_{\mathcal{A}}(x)}{\sum_{i=1}^{4} P(c^i, e) + UB_{\mathcal{A}}(x) + LB_{\mathcal{A}}(x')}$$

From Eq. (15) and (16) we get:

$$P_{\mathcal{A}}^L(x|e) = \frac{\sum_{i=1}^{4} P(x, c^i, e) + LB_{\mathcal{A}}(x)}{\sum_{i=1}^{4} P(c^i, e) + LB_{\mathcal{A}}(x) + UB_{\mathcal{A}}(x')}$$

The total number of tuples processed is $M' = 4 + 4 = 8 < 24$.

## 3.4 $ATB$ Properties

In this section we analyze the time complexity of the $ATB$ framework, evaluate its worst-case lower and upper bounds, and analyze the monotonicity properties of its bounds interval (as a function of $h$).

THEOREM 3.1 (complexity) *Given an algorithm $\mathcal{A}$ that computes lower and upper bounds on joint probabilities $P(c_{1:q_i}, e)$ and $P(x, c_{1:q_i}, e)$ in time $O(T)$, and a loop-cutset $C$, $P_{\mathcal{A}}^L(x|e)$ and $P_{\mathcal{A}}^U(x|e)$ are computed in time $O(h \cdot N + T \cdot h \cdot (d-1) \cdot |C|)$ where $d$ is the maximum domain size and $N$ is the problem input size.*

PROOF. Since $C$ is a loop-cutset, the exact probabilities $P(c^i, e)$ and $P(x, c^i, e)$ can be computed in time $O(N)$. From Proposition 3.1, there are $O(h \cdot (d-1) \cdot |C|)$ partially-instantiated tuples. Since algorithm $\mathcal{A}$ computes upper and lower bounds on $P(c_{1:q_j}^j, e)$ and $P(x, c_{1:q_j}^j, e)$ in time $O(T)$, the bounds on partially-instantiated tuples can be computed in time $O(T \cdot h \cdot (d-1) \cdot |C|)$. $\square$

Let the plug-in algorithm $\mathcal{A}$ be a brute-force algorithm, denoted $BF$, that trivially instantiates $P_{BF}^L(x, c_{1:q_j}^j, e) = 0$, $P_{BF}^U(x, c_{1:q_j}^j, e) = P(c_{1:q_j}^j)$, and $UB[P(c_{1:q_j}^j, e]] = P(c_{1:q_j}^j)$. Then, from Eq. (15):

$$P_{BF}^L(x|e) \triangleq \frac{\sum_{i=1}^h P(x, c^i, e)}{\sum_{i=1}^h P(c^i, e) + \sum_{j=1}^{M'} P(c_{1:q_j}^j)} = \frac{\sum_{i=1}^h P(x, c^i, e)}{\sum_{i=1}^h P(c^i, e) + \sum_{j=h+1}^M P(c^j)} \tag{21}$$

while from Eq. (19):

$$P_{BF}^U(x|e) \triangleq \frac{\sum_{i=1}^h P(x, c^i, e) + \sum_{j=1}^{M'} P(c_{1:q_j}^j)}{\sum_{i=1}^h P(c^i, e) + \sum_{j=1}^{M'} P(c_{1:q_j}^j)} = \frac{\sum_{i=1}^h P(x, c^i, e) + \sum_{j=h+1}^M P(c^j)}{\sum_{i=1}^h P(c^i, e) + \sum_{j=h+1}^M P(c^j)} \tag{22}$$

Assuming that algorithm $\mathcal{A}$ computes bounds at least as good as those computed by $BF$, $P_{BF}^L(x|e)$ and $P_{BF}^U(x|e)$ are the worst-case bounds computed by $ATB$.

Now, we are ready to compute an upper bound on the $ATB$ bounds interval:

THEOREM **3.2** ($ATB$ **bounds interval upper bound**) $ATB$ *length of the interval between its lower and upper bounds is upper bounded by a monotonic non-increasing function of $h$:*

$$P_{\mathcal{A}}^U(x|e) - P_{\mathcal{A}}^L(x|e) \leq \frac{\sum_{j=h+1}^M P(c^j)}{\sum_{i=1}^h P(c^i, e) + \sum_{j=h+1}^M P(c^j)} \triangleq I_h$$

PROOF. See Appendix C. □

Next we show that $ATB$ lower and upper bounds are as good or better than the bounds computed by $BC$.

THEOREM **3.3 (tighter lower bound)** $P_{\mathcal{A}}^L(x|e) \geq P_{BC}^L(x|e)$.

PROOF. $P_{BF}^L(x|e)$ is the worst-case lower bound computed by $ATB$. Since $P_{BF}^L(x|e) = P_{BC}^L(x|e)$, and $P_{\mathcal{A}}^L(x|e) \geq P_{BF}^L(x|e)$, then $P_{\mathcal{A}}^L(x|e) \geq P_{BC}^L(x|e)$. □

THEOREM **3.4 (tighter upper bound)** $P_{\mathcal{A}}^U(x|e) \leq P_{BC}^U(x|e)$.

PROOF. $P_{BF}^U(x|e)$ is the worst-case upper bound computed by $ATB$. Since $P_{BF}^U(x|e) \leq P_{BC}^U(x|e)$ due to lemma 3.1, it follows that $P_{\mathcal{A}}^U(x|e) \leq P_{BC}^U(x|e)$. □

## 4. Experimental Evaluation

The purpose of the experiments is to evaluate the performance of our $ATB$ framework on the two probabilistic tasks of single-variable posterior marginals and probability of evidence. The experiments on the first task were conducted on 1.8Ghz CPU with 512 MB RAM, while the experiments on the second task were conducted on 2.66GHz CPU with 2GB RAM.

Recall that $ATB$ has a control parameter $h$ that fixes the number of cutset tuples for which the algorithm computes its exact joint probability. Given a fixed $h$, the quality of the bounds will presumably depend on the ability to select $h$ high probability cutset tuples. In our implementation, we use an optimized version of Gibbs sampling, that during the sampling process maintains a list of the $h$ tuples having the highest joint probability. As noted, other schemes should be considered for this subtask as part of the future work. We obtain the loop-cutset using $mga$ algorithm (Becker & Geiger, 1996).

Before we report the results, we describe bound propagation and its variants, which we use as a plug-in algorithm $\mathcal{A}$ and also as a stand-alone bounding scheme.

## 4.1 Bound Propagation

*Bound propagation* $(BdP)$ (Leisink & Kappen, 2003) is an iterative algorithm that bounds the posterior marginals of a variable. The bounds are initialized to 0 and 1 and are iteratively improved by solving a linear optimization problem for each variable $X \in \mathcal{X}$ such that the minimum and maximum of the objective function correspond to the lower and upper bound on the posterior marginal $P(x|e)$, $x \in \mathcal{D}(X)$.

We cannot directly plug $BdP$ into $ATB$ to bound $P(c_{1:q}, e)$ because it only bounds conditional probabilities. Thus, we factorize $P(c_{1:q}, e)$ as follows:

$$P(c_{1:q}, e) = \prod_{e_j \in E} P(e_j | e_1, \ldots, e_{j-1}, c_{1:q}) P(c_{1:q})$$

Each factor $P(e_j | e_1, \ldots, e_{j-1}, c_{1:q})$ can be bounded by $BdP$, while $P(c_{1:q})$ can be computed exactly since the relevant subnetwork over $c_{1:q}$ (see Def. 2.5) is singly connected. Let $P_{BdP}^L$ and $P_{BdP}^U$ denote the lower and upper bounds computed by $BdP$ on some marginal. The bounds $BdP$ computes on the joint probability are:

$$\prod_{e_j \in E} P_{BdP}^L(e_j | e_1, \ldots, e_{j-1}, c_{1:q}) P(c_{1:q}) \leq P(c_{1:q}, e) \leq \prod_{e_j \in E} P_{BdP}^U(e_j | e_1, \ldots, e_{j-1}, c_{1:q}) P(c_{1:q})$$

Note that $BdP$ has to bound a large number of tuples when plugged into $ATB$, and therefore, solve a large number of linear optimization problems. The number of variables in each problem is exponential in the size of the Markov blanket of $X$.

As a baseline for comparison with $ATB$, we use in our experiments a variant of bound propagation called $BdP+$ (Bidyuk & Dechter, 2006b) that exploits the structure of the network to restrict the computation of $P(x|e)$ to the relevant subnetwork of $X$ (see Def. 2.5). The Markov boundary of $X$ (see Def. 2.4) within relevant subnetwork of $X$ does not include the children of $X$ that are not observed and have no observed descedants; therefore, it is a subnetwork of the Markov boundary in the original network. Sometimes, the Markov boundary of $X$ is still too big to compute under limited memory resouces. $BdP+$ uses a parameter $k$ to specify the maximum size of the Markov boundary domain space. The algorithm skips the variables whose Markov boundary domain size exceeds $k$, and so their lower and upper bound values remain 0 and 1, respectively. When some variables are skipped, the bounds computed by $BdP+$ for the remaining variables may be less accurate.

| network | N | w* | |LC| | $|\mathcal{D}(\mathbf{LC})|$ | $|E|$ | Time(BE) | Time(LC) |
|---------|---|----|----|-----|-----|----------|----------|
| Alarm | 37 | 4 | 5 | 108 | 1-4 | 0.01 sec | 0.05 sec |
| Barley | 48 | 7 | 12 | $> 2^{27}$ | 4-8 | 50 sec | >22 hrs[1] |
| cpcs54 | 54 | 15 | 15 | 32768 | 2-8 | 1 sec | 22 sec |
| cpcs179 | 179 | 8 | 8 | 49152 | 12-24 | 2 sec | 37 sec |
| cpcs360b | 360 | 21 | 26 | $2^6$ | 11-23 | 20 min | $> 8$ hrs[1] |
| cpcs422b | 422 | 22 | 47 | $2^7$ | 4-10 | 50 min | $> 2 \times 10^9$ hrs[1] |
| Munin3 | 1044 | 7 | 30 | $> 2^{30}$ | 257 | 8 sec | $> 1700$ hrs[1] |
| Munin4 | 1041 | 8 | 49 | $> 2^{49}$ | 235 | 70 sec | $> 1 \times 10^8$ hrs[1] |

Table 1: Complexity characteristics of the benchmarks from the UAI repository: $N$-number of nodes, $w^*$-induced width, $|LC|$-number of nodes in a loop-cutset, $|\mathcal{D}(LC)|$-loop-cutset state space size, Time(BE) is the exact computation time via bucket elimination, Time(LC) is the exact computation time via loop-cutset conditioning. The results are averaged over a set of network instances with different evidence. Evidence nodes and their values are selected at random.

Our preliminary tests showed that plugging $BdP+$ into $ATB$ is timewise infeasible (even for small $k$). Instead, we developed and used a different version of bound propagation called $ABdP+$ (Bidyuk & Dechter, 2006b) as a plug-in algorithm $\mathcal{A}$, which was more cost-effective in terms of accuracy and time overhead. $ABdP+$ includes the same enhancements as $BdP+$, but solves the linear optimization problem for each variable using an approximation algorithm. This implies that we obtain bounds faster but they are not as accurate. Roughly, the relaxed linear optimization problem can be described as a fractional packing and covering with multiple knapsacks and solved by a fast greedy algorithm (Bidyuk & Dechter, 2006b). $ABdP+$ is also parameterized by $k$ to control the maximum size of the linear optimization problem. Thus, $ATB$ using $ABdP+$ as a plug-in has two control parameters: $h$ and $k$.

## 4.2 Bounding Single-Variable Marginals

We compare the performance of the following three algorithms: $ATB$ (with $ABdP+$ as a plug-in), $BdP+$, as described in the previous section, and $BBdP+$ (Bidyuk & Dechter, 2006a). The latter is a combination of $ATB$ and $BdP+$. First, we run algorithm $ATB$ with $ABdP+$ plug-in. Then, we use the bounds computed by $ATB$ to initialize bounds in $BdP+$ (instead of 0 and 1) and run $BdP+$. Note that, given fixed values of $h$ and $k$, $BBdP+$ will always compute tighter bounds than either $ATB$ and $BdP+$. Our goal is to analyze its trade-off between the increase of the bounds' accuracy and the computation time overhead. We also compare with *approximate decomposition* ($AD$) (Larkin, 2003) whenever it is feasible and relevant. We did not include the results for the stand-alone $ABdP+$ since our objective was to compare $ATB$ bounds with the most accurate bounds obtained by bound propagation. Bidyuk (2006) provides additional comparison with various refinements of $BdP$ (Bidyuk & Dechter, 2006b) mentioned earlier.

---

1. Times are extrapolated.

### 4.2.1 Benchmarks

We tested our framework on four different benchmarks: *Alarm*, *Barley*, *CPCS*, and *Munin*. *Alarm network* is a model for monitoring patients undergoing surgery in an operating room (Beinlich, Suermondt, Chavez, & Cooper, 1989). *Barley network* is a part of the decision-support system for growing malting barley (Kristensen & Rasmussen, 2002). *CPCS networks* are derived from the Computer-Based Patient Care Simulation system and based on INTERNIST-1 and Quick Medical Reference Expert systems (Pradhan, Provan, Middleton, & Henrion, 1994). We experiment with cpcs54, cpcs179, cpcs360b, and cpcs422b networks. *Munin networks* are a part of the expert system for computer-aided electromyography (Andreassen, Jensen, Andersen, Falck, Kjærulff, Woldbye, Srensen, Rosenfalck, & Jensen, 1990). We experiment with Munin3 and Munin4 networks. For each network, we generated 20 different sets of evidence variables picked at random. For Barley network, we select evidence variables as defined by Kristensen and Rasmussen (2002).

Table 1 summarizes the characteristic of each network. For each one, the table specifies the number of variables $N$, the induced width $w^*$, the size of loop cutset $|LC|$, the number of loop-cutset tuples $|\mathcal{D}(LC)|$, and the time needed to compute the exact posterior marginals by bucket-tree elimination (exponential in the induced width $w^*$) and by cutset conditioning (exponential in the size of loop-cutset).

Computing the posterior marginals exactly is easy in Alarm network, cpcs54, and cpcs179 using either bucket elimination or cutset conditioning since they have small induced width and a small loop-cutset. We include those benchmarks as a proof of concept only. Several other networks, Barley, Munin3, and Munin4, also have small induced width and, hence, their exact posterior marginals can be obtained by bucket elimination. However, since $ATB$ is linear in space, it should be compared against linear-space schemes such as cutset-conditioning. From this perspective, Barley, Munin3, and Munin4 are hard. For example, Barley network has only 48 variables, its induced width is $w^* = 7$, and exact inference by bucket elimination takes only 30 seconds. Its loop-cutset contains only 12 variables, but the number of loop-cutset tuples exceeds 2 million because some variables have large domain sizes (up to 67 values). Enumerating and computing all cutset tuples, at a rate of about 1000 tuples per second, would take over 22 hours. Similar considerations apply in case of Munin3 and Munin4 networks.

### 4.2.2 Measures of Performance

We measure the quality of the bounds via the average length of the interval between lower and upper bound:

$$\overline{I} = \frac{\sum_{X \in \mathcal{X}} \sum_{x \in \mathcal{D}(X)} (P^U(x|e) - P^L(x|e))}{\sum_{X \in \mathcal{X}} |\mathcal{D}(X)|}$$

We approximate posterior marginal as the midpoint between lower and upper bound in order to show whether the bounds are well-centered around the posterior marginal $P(x|e)$. Namely:

$$\hat{P}(x|e) = \frac{P_{\mathcal{A}}^U(x|e) + P_{\mathcal{A}}^L(x|e)}{2}$$

and then measure the average absolute error $\Delta$ with respect to that approximation:

$$\Delta = \frac{\sum_{X \in \mathcal{X}} \sum_{x \in \mathcal{D}(X)} |P(x|e) - \hat{P}(x|e)|}{\sum_{X \in \mathcal{X}} |\mathcal{D}(X)|}$$

Finally, we report $\%P(e) = \frac{\sum_{i=1}^{h} P(x, c^i, e)}{P(e)} \times 100\%$ that was covered by the explored cutset tuples. Notably, in some benchmarks, a few thousand cutset tuples is enough to cover $> 90\%$ of $P(e)$.

### 4.2.3 Results

We summarize the results for each benchmark in a tabular format and charts. We highlight in bold face the first $ATB$ data point where the average bounds interval is as good or better than $BdP+$. The charts show the convergence of the bounds interval length as a function of $h$ and time.

For $ATB$ and $BBdP+$ the maximum Markov boundary domain size was fixed at $k = 2^{10}$. For $BdP+$, we vary parameter $k$ from $2^{14}$ to $2^{19}$. Note that $BdP+$ only depends on $k$, not on $h$. In the tables, we report the best result obtained by $BdP+$ and its computation time so that it appears as constant with respect to $h$. However, when we plot accuracy against time, we include $BdP+$ bounds obtained using smaller values of parameter $k$. In the case of Alarm network, varying $k$ did not make any difference since the full Markov boundary domain size equals $2^{10} < 2^{14}$. The computation time of $BBdP+$ includes the $ATB$ plus the $BdP+$ time.

**Alarm network.** Figure 3 reports the results. Since the maximum Markov boundary in Alarm network is small, $BdP+$ runs without limitations and computes an average bounds interval of 0.61 in 4.3 seconds. Note that the enumeration of less than the 25% of the total number of cutset tuples covers 99% of the $P(e)$. This fact suggests that schemes based on cutset conditioning should be very suitable for this benchmark. Indeed, $ATB$ outperforms $BdP+$, computing more accurate bounds starting with the first data point of $h = 25$ where the mean interval $\overline{I}_{ATB} = 0.41$ while the computation time is 0.038 seconds, an order of magnitude less than $BdP+$. The extreme efficiency of $ATB$ in terms of time is clearly seen in the right chart. The x-axis scale is logarithmic to fit all the results. As expected, the average bounds interval generated by $ATB$ and $BBdP+$ decrease as the number of cutset tuples $h$ increases, demonstrating the anytime property of $ATB$ with respect to $h$. Given a fixed $h$, $BBdP+$ has a very significant overhead in time with respect to $ATB$ (two orders of magnitude for values of $h$ smaller than 54) and only a minor improvement in accuracy.

**Barley network**. Figure 4 reports the results. $ATB$ and $BBdP+$ improve as $h$ increases. However, the improvement is quite moderate while very time consuming due to more uniform shape of the distribution $P(C|e)$ as reflected by the very small % of $P(e)$ covered by explored tuples (only 1% for 562 tuples and only 52% for 12478 tuples). For example, the average $ATB$ (resp. $BBdP+$) bounds interval decreases from 0.279 (resp. 0.167), obtained in 9 (resp. 10) seconds, to 0.219 (resp. 0.142) obtained in 139 (resp. 141) seconds. Given a fixed $h$, $BBdP+$ substantially improves $ATB$ bounds with little time overhead (2 seconds in general). Namely, in this benchmark, $BBdP+$ computation time is dominated by $ATB$ computation time. Note that the computation time of the stand-alone $BdP+$ algorithm is less than 2 seconds. Within that time, $BdP+$ yields an average interval length of 0.23, while $ATB$ and $BBdP+$ spend 86 and 10 seconds, respectively, to obtain the same quality bounds. However, the anytime behavior of the latter algorithms allows them to improve with time, a very desirable characteristic when computing bounds. Moreover, note that its overhead in time with respect to $ATB$ is completely negligible.

| Alarm, N=37, $w^*$=5, $|LC|$=8, $|D_{LC}|$=108, $|E|$=1-4 | | | | | | | | | | |
|---|---|---|---|---|---|---|---|---|---|---|
| | | *BdP+* | | | | *ATB* | | | *BBdP+* | |
| h | %P(e) | $\overline{I}$ | $\Delta$ | *time(sec)* | $\overline{I}$ | $\Delta$ | *time(sec)* | $\overline{I}$ | $\Delta$ | *time(sec)* |
| **25** | 86 | 0.61 | 0.21 | *4.3* | **0.41** | 0.12 | **0.038** | 0.35 | 0.10 | *3.4* |
| 34 | 93 | 0.61 | 0.21 | *4.3* | 0.31 | 0.09 | *0.039* | 0.27 | 0.08 | *2.3* |
| 40 | 96 | 0.61 | 0.21 | *4.3* | 0.25 | 0.07 | *0.044* | 0.22 | 0.06 | *2.1* |
| 48 | 97 | 0.61 | 0.21 | *4.3* | 0.24 | 0.05 | *0.051* | 0.15 | 0.04 | *1.5* |
| 50 | 98 | 0.61 | 0.21 | *4.3* | 0.16 | 0.04 | *0.052* | 0.12 | 0.03 | *1.2* |
| 54 | 99 | 0.61 | 0.21 | *4.3* | 0.13 | 0.03 | *0.059* | 0.09 | 0.02 | *0.86* |

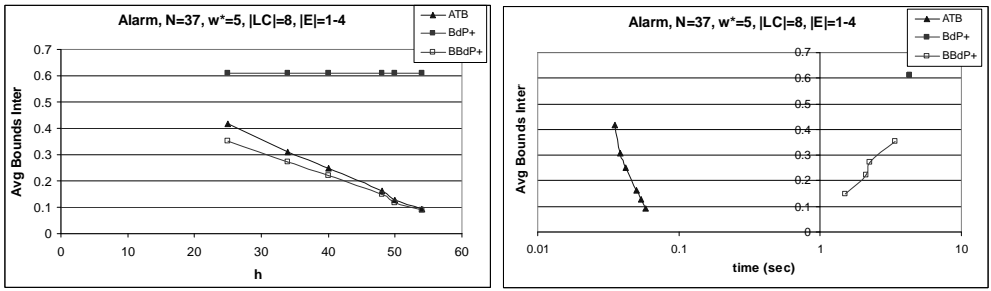

Figure 3: Results for Alarm network. The table reports the average bounds interval $\overline{I}$, average error $\Delta$, computation time (in seconds), and percent of probability of evidence $P(e)$ covered by the fully-instantiated cutset tuples as a function of $h$. We highlight in bold face the first $ATB$ data point where the average bounds interval is as good or better than $BdP+$. The charts show the convergence of the bounds interval length as a function of $h$ and time. $BdP+$ uses full size Markov boundary since its domain size is small ($< 2^{14}$), resulting in only one data point on the chart on the right.

**CPCS networks.** Figures 5 to 8 show the results for cpcs54, cpcs179, cpcs360b and cpcs422b, respectively. The behavior of the algorithms in all networks is very similar. As in the previous benchmarks, $ATB$ and $BBdP+$ bounds interval decreases as $h$ increases. Given a fixed $h$, $BBdP+$ computes slightly better bounds intervals than $ATB$ in all networks but cpcs179. For all networks, $BBdP+$ has overhead in time with respect to $ATB$. This overhead is constant for all values of $h$ and for all networks except for cpcs54, for which the overhead decreases as $h$ increases. $ATB$ and $BBdP+$ outperform $BdP+$. Both algorithms compute the same bound interval length as $BdP+$, improving the computation time in one order of magnitude. Consider for example cpcs422b, a challenging instance for any inference scheme as it has relatively large induced width and loop-cutset size. $ATB$ outperforms $BdP+$ after 50 seconds starting with $h = 1181$, and $BBdP+$ outperforms $BdP+$ in 37 seconds starting with $h = 253$ ($BdP+$ convergence is shown in the plot, but only the best result is reported in the table).

| Barley, $N$=48, $w^*$=7, $|LC|$=12, $|D_{LC}| > 2 \times 10^6$, $|E|$=4-8 | | | | | | | | | | |
|---|---|---|---|---|---|---|---|---|---|---|
| | | BdP+ | | | ATB | | | BBdP+ | | |
| h | %P(e) | $\overline{I}$ | $\Delta$ | time(sec) | $\overline{I}$ | $\Delta$ | time(sec) | $\overline{I}$ | $\Delta$ | time(sec) |
| 562 | 1 | 0.23 | 0.07 | 1.7 | 0.279 | 0.097 | 9 | 0.167 | 0.047 | 10 |
| 1394 | 3 | 0.23 | 0.07 | 1.7 | 0.263 | 0.090 | 23 | 0.162 | 0.045 | 25 |
| 2722 | 6 | 0.23 | 0.07 | 1.7 | 0.247 | 0.084 | 43 | 0.154 | 0.042 | 45 |
| 4429 | 14 | 0.23 | 0.07 | 1.7 | 0.235 | 0.079 | 65 | 0.147 | 0.040 | 67 |
| **6016** | 22 | 0.23 | 0.07 | 1.7 | **0.230** | 0.078 | 86 | 0.145 | 0.040 | 88 |
| 7950 | 33 | 0.23 | 0.07 | 1.7 | 0.228 | 0.077 | 99 | 0.145 | 0.040 | 101 |
| 9297 | 40 | 0.23 | 0.07 | 1.7 | 0.224 | 0.075 | 111 | 0.143 | 0.039 | 113 |
| 12478 | 52 | 0.23 | 0.07 | 1.7 | 0.219 | 0.073 | 139 | 0.142 | 0.038 | 141 |

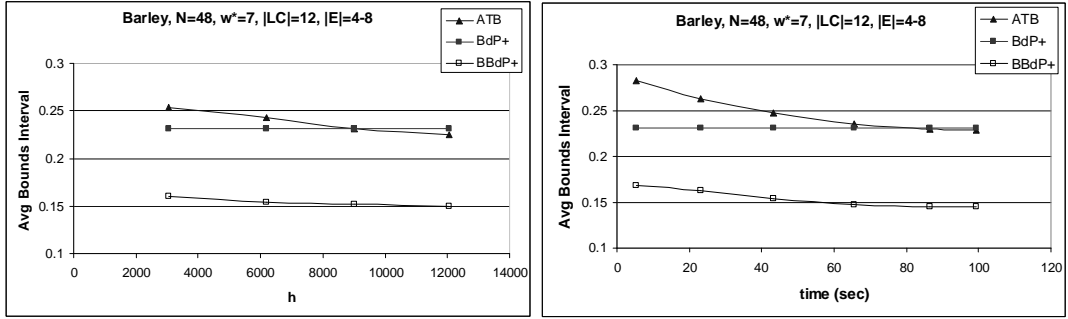

Figure 4: Results for Barley network. The table reports the average bounds interval $\overline{I}$, average error $\Delta$, computation time (in seconds), and percent of probability of evidence $P(e)$ covered by the fully-instantiated cutset tuples as a function of $h$. We highlight in bold face the first $ATB$ data point where the average bounds interval is as good or better than $BdP+$. The charts show the convergence of the bounds interval length as a function of $h$ and time.

Larkin (2003) reported bounds on cpcs360b and cpcs422b using $AD$ algorithm. For the first network, $AD$ achieved bounds interval length of 0.03 in 10 seconds. Within the same time, $ATB$ computes an average bounds interval of $\approx 0.005$. For cpcs422b, $AD$ achieved bounds interval of 0.15, obtained in 30 seconds. Within the same time, $ATB$ and $BBdP+$ obtain comparable results computing average bounds interval of 0.24 and 0.15, respectively. It is important to note that the comparison is not on the same instances since the evidence nodes are not the same. Larkin's code was not available for further experiments.

**Munin networks.** Figure 9 reports the results for both Munin networks. Let us first consider Munin3 network. Given a fixed $h$, $ATB$ and $BBdP+$ compute almost identical bound intervals with $BBdP+$ having a noticeable time overhead. Note that the two curves in the chart showing convergence as a function of $h$ are very close and hard to distinguish, while the points of $BBdP+$ in the chart showing convergence as a function of time are shifted to the right with respect to the ones of $ATB$. $ATB$ is clearly superior to $BdP+$ both in accuracy and time. $BdP+$ computes bounds interval of 0.24 within 12 seconds, while $ATB$ computes bounds interval of 0.050 in 8 seconds. In Munin4, given a fixed $h$, $BBdP+$ computes tighter bounds than $ATB$ with some time overhead. However, the improvement decreases as $h$ increases as shown by the convergence of both curves either as a function of $h$ and time. Since the loop-cutset size is large, the convergence of $ATB$ is relatively slow. $BdP+$ computes bounds interval of 0.23 within 15 seconds, while $ATB$ and $BBdP+$ compute bounds of the same quality within 54 and 21 seconds, respectively.

| cpcs54, $N$=54, $|LC|$=15, $w^*$=15, $|D_{LC}|$=32678, $|E|$=2-8 | | | | | | | | | | |
|---|---|---|---|---|---|---|---|---|---|---|
| | | BdP+ | | | ATB | | | BBdP+ | | |
| h | %P(e) | $\bar{I}$ | $\Delta$ | $time(sec)$ | $\bar{I}$ | $\Delta$ | $time(sec)$ | $\bar{I}$ | $\Delta$ | $time(sec)$ |
| 513 | 10 | 0.35 | 0.02 | 24 | 0.51 | 0.027 | 0.9 | 0.34 | 0.011 | 3.1 |
| 1114 | 19 | 0.35 | 0.02 | 24 | 0.45 | 0.023 | 1.5 | 0.32 | 0.010 | 3.1 |
| 1581 | 29 | 0.35 | 0.02 | 24 | 0.42 | 0.021 | 1.9 | 0.31 | 0.009 | 3.4 |
| 1933 | 34 | 0.35 | 0.02 | 24 | 0.40 | 0.020 | 2.2 | 0.30 | 0.009 | 3.6 |
| 2290 | 40 | 0.35 | 0.02 | 24 | 0.38 | 0.019 | 2.4 | 0.30 | 0.008 | 3.9 |
| 2609 | 46 | 0.35 | 0.02 | 24 | 0.37 | 0.018 | 2.7 | 0.29 | 0.007 | 4.0 |
| **3219** | **53** | **0.35** | **0.02** | **24** | **0.34** | **0.016** | **3.2** | **0.27** | **0.007** | **4.5** |
| 3926 | 59 | 0.35 | 0.02 | 24 | 0.31 | 0.014 | 3.8 | 0.25 | 0.006 | 5.2 |
| 6199 | 63 | 0.35 | 0.02 | 24 | 0.23 | 0.010 | 5.9 | 0.20 | 0.006 | 6.6 |
| 7274 | 68 | 0.35 | 0.02 | 24 | 0.20 | 0.008 | 6.9 | 0.17 | 0.006 | 7.3 |

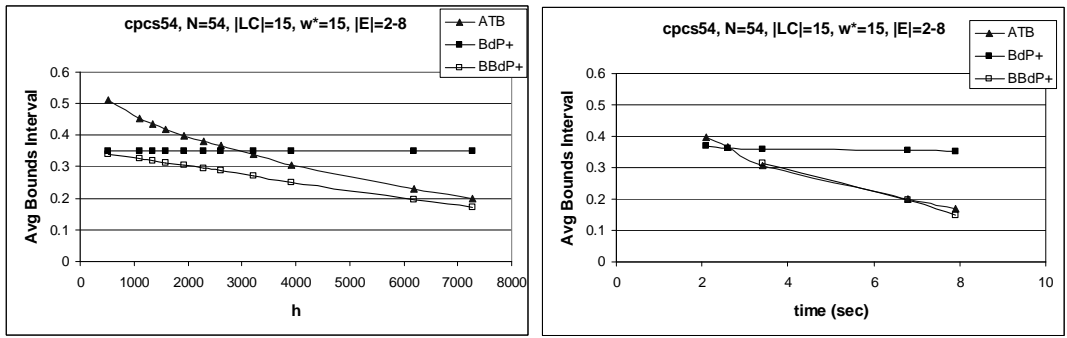

Figure 5: Results for cpcs54 network. The table reports the average bounds interval $\bar{I}$, average error $\Delta$, computation time (in seconds), and percent of probability of evidence $P(e)$ covered by the fully-instantiated cutset tuples as a function of $h$. We highlight in bold face the first $ATB$ data point where the average bounds interval is as good or better than $BdP+$. The charts show the convergence of the bounds interval length as a function of $h$ and time.

## 4.3 Bounding the Probability of Evidence

We compare the performance of the following three algorithms: $ATB$, *mini-bucket elimination* ($MBE$) (Dechter & Rish, 2003), and *variable elimination and conditioning* ($VEC$). For $ATB$, we test different configurations of the control parameters $(h, k)$. Note that when $h = 0$, $ATB$ is equivalent to its plug-in algorithm $\mathcal{A}$, which in our case is $ABdP+$.

### 4.3.1 Algorithms and Benchmarks

$MBE$ is a general bounding algorithm for graphical model problems. In particular, given a Bayesian network, $MBE$ computes lower and upper bound on the probability of evidence.

| cpcs179, $N$=179, $w^*$=8, $|LC|$=8, $|D_{LC}|$=49152, $|E|$=12-24 | | | | | | | | | | | |
|---|---|---|---|---|---|---|---|---|---|---|---|
| | | $BdP+$ | | | $ATB$ | | | $BBdP+$ | | | |
| h | %P(e) | $\bar{I}$ | $\Delta$ | time(sec) | $\bar{I}$ | $\Delta$ | time(sec) | $\bar{I}$ | $\Delta$ | time(sec) | |
| 242 | 70 | 0.15 | 0.05 | 20 | 0.22 | 0.067 | 4 | 0.092 | 0.029 | 11 | |
| **334** | 75 | 0.15 | 0.05 | 20 | **0.12** | **0.033** | **6** | 0.054 | 0.016 | 13 | |
| 406 | 78 | 0.15 | 0.05 | 20 | 0.09 | 0.024 | 7 | 0.037 | 0.010 | 13 | |
| 574 | 82 | 0.15 | 0.05 | 20 | 0.07 | 0.018 | 9 | 0.029 | 0.008 | 15 | |
| 801 | 85 | 0.15 | 0.05 | 20 | 0.05 | 0.014 | 10 | 0.022 | 0.006 | 17 | |
| 996 | 87 | 0.15 | 0.05 | 20 | 0.04 | 0.010 | 12 | 0.017 | 0.005 | 18 | |
| 1285 | 88 | 0.15 | 0.05 | 20 | 0.03 | 0.006 | 13 | 0.012 | 0.003 | 20 | |
| 1669 | 90 | 0.15 | 0.05 | 20 | 0.02 | 0.003 | 16 | 0.007 | 0.002 | 22 | |

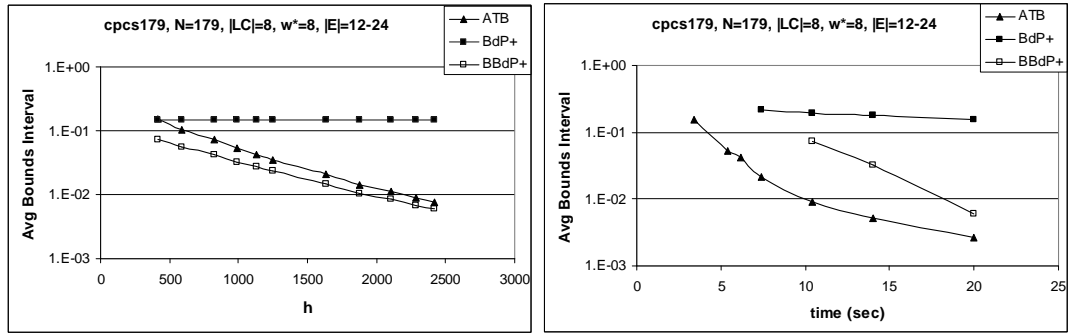

Figure 6: Results for cpcs179 network. The table reports the average bounds interval $\bar{I}$, average error $\Delta$, computation time (in seconds), and percent of probability of evidence $P(e)$ covered by the fully-instantiated cutset tuples as a function of $h$. We highlight in bold face the first $ATB$ data point where the average bounds interval is as good or better than $BdP+$. The charts show the convergence of the bounds interval length as a function of $h$ and time.

$MBE$ has a control parameter $z$, that allows trading time and space for accuracy. As the value of the control parameter $z$ increases, the algorithm computes tighter bounds using more time and space, which is exponential in $z$.

$VEC$ is an algorithm that combines conditioning and variable elimination. It is based on the $w$-cutset conditioning scheme. Namely, the algorithm conditions or instantiates enough variables so that the remaining problem conditioned on the instantiated variables can be solved exactly using *bucket elimination* (Dechter, 1999). The exact probability of evidence can be computed by summing over the exact solution output by bucket elimination for all possible instantiations of the $w$-cutset. When $VEC$ is terminated before completion, it outputs a partial sum yielding a lower bound on the probability of evidence. The implementation of $VEC$ is publicly available[1].

We tested $ATB$ for bounding $P(e)$ on three different benchmarks: *Two-layer Noisy-Or*, *grids* and *coding* networks. All instances are included in the UAI08 evaluation[2].

In *two-layer noisy-or networks*, variables are organized in two layers where the ones in the second layer have 10 parents. Each probability table represents a noisy OR-function.

---

1. http://graphmod.ics.uci.edu/group/Software

2. http://graphmod.ics.uci.edu/uai08/Evaluation/Report

| cpcs360b, N=360, $w^*$ = 21, $|LC|$ = 26, $|D_{LC}|$=$2^{26}$, $|E|$=11-23 | | | | | | | | | | |
|---|---|---|---|---|---|---|---|---|---|---|
| | | $BdP+$ | | | $ATB$ | | | BBdP+ | | |
| h | %P(e) | $\overline{I}$ | $\Delta$ | $time(sec)$ | $\overline{I}$ | $\Delta$ | $time(sec)$ | $\overline{I}$ | $\Delta$ | $time(sec)$ |
| 121 | 83 | 0.027 | 0.009 | $55$ | 0.0486 | 1.6E-2 | $5$ | 0.0274 | 1.0E-2 | $7$ |
| **282** | 92 | 0.027 | 0.009 | $55$ | **0.0046** | 9.0E-4 | **10** | 0.0032 | 8.5E-4 | $12$ |
| 501 | 96 | 0.027 | 0.009 | $55$ | 0.0020 | 3.6E-4 | $15$ | 0.0014 | 3.5E-4 | $17$ |
| 722 | 97 | 0.027 | 0.009 | $55$ | 0.0012 | 2.4E-4 | $19$ | 0.0009 | 2.3E-4 | $21$ |
| 938 | 98 | 0.027 | 0.009 | $55$ | 0.0006 | 8.4E-5 | $25$ | 0.0004 | 7.8E-5 | $27$ |
| 1168 | 98 | 0.027 | 0.009 | $55$ | 0.0005 | 7.5E-5 | $29$ | 0.0004 | 6.9E-5 | $31$ |
| 1388 | 99 | 0.027 | 0.009 | $55$ | 0.0004 | 5.9E-5 | $35$ | 0.0003 | 5.4E-5 | $37$ |
| 1582 | 99 | 0.027 | 0.009 | $55$ | 0.0003 | 5.3E-5 | $39$ | 0.0002 | 4.8E-5 | $41$ |
| 1757 | 99 | 0.027 | 0.009 | $55$ | 0.0003 | 4.7E-5 | $43$ | 0.0002 | 4.4E-5 | $46$ |

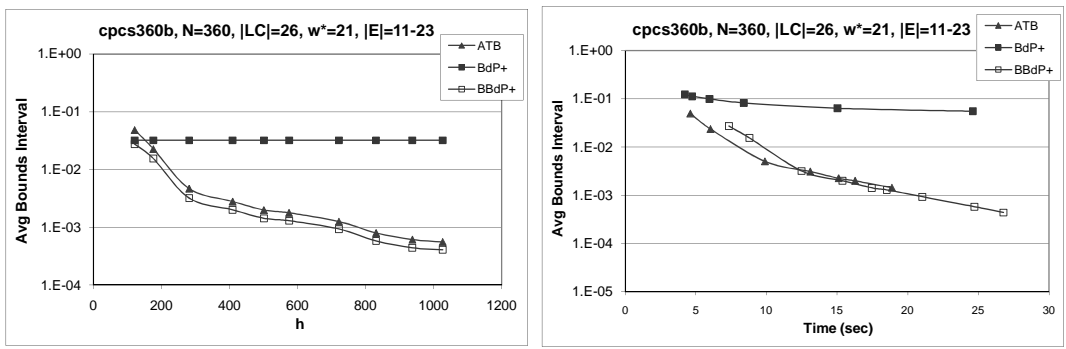

Figure 7: Results for cpcs360b. The table reports the average bounds interval $\overline{I}$, average error $\Delta$, computation time (in seconds), and percent of probability of evidence $P(e)$ covered by the fully-instantiated cutset tuples as a function of $h$. We highlight in bold face the first $ATB$ data point where the average bounds interval is as good or better than $BdP+$. The charts show the convergence of the bounds interval length as a function of $h$ and time.

Each parent variable $y_j$ has a value $P_j \in [0..P_{noise}]$. The CPT for each variable in the second layer is then defined as $P(x = 0 | y_1, \ldots, y_P) = \prod_{y_j=1} P_j$ and $P(x = 1 | y_1, \ldots, y_P) = 1 - P(x = 0 | y_1, \ldots, y_P)$. We experiment with a class of problems called *bn2o* instances in the UAI08.

In *grid networks*, variables are organized as an $M \times M$ grid. We experiment with *grids2* instances, as they were called in UAI08, which are characterized by two parameters $(M, D)$, where $D$ is the percentage of determinism (i.e., the percentage of values in all CPTs assigned to either 0 or 1). For each parameter configuration, 10 samples were generated by randomly assigning value 1 to one leaf node. In UAI08 competition, these instances were named $D$-$M$-$I$, where $I$ is the instance number.

*Coding networks* can be represented as a four layer Bayesian network having $M$ nodes in each layer. The second and third layer correspond to input information bits and parity check bits respectively. Each parity check bit represents an XOR function of input bits. Input and parity check nodes are binary while the output nodes are real-valued. We consider the $BN\_126$ to $BN\_134$ instances in the UAI08 evaluation. Each one has $M = 128$, 4 parents for each node and channel noise variance ($\sigma = 0.40$). These networks are very hard and exact results are not available.

| cpcs422b, N=422, $w^*$ = 22, $|LC|$ = 47, $|D_{LC}|$=$2^{47}$, $|E|$=4-10 | | | | | | | | | | |
|---|---|---|---|---|---|---|---|---|---|---|
| | | BdP+ | | | | ATB | | | BBdP+ | | |
| h | %P(e) | $\overline{I}$ | $\Delta$ | time(sec) | $\overline{I}$ | $\Delta$ | time(sec) | $\overline{I}$ | $\Delta$ | time(sec) |
| 64 | 1.7 | 0.19 | 0.06 | 120 | 0.28 | 0.100 | 21 | 0.19 | 0.056 | 23 |
| 256 | 2.0 | 0.19 | 0.06 | 120 | 0.24 | 0.090 | 26 | 0.15 | 0.050 | 35 |
| 379 | 2.6 | 0.19 | 0.06 | 120 | 0.22 | 0.078 | 32 | 0.14 | 0.049 | 41 |
| 561 | 2.9 | 0.19 | 0.06 | 120 | 0.20 | 0.073 | 36 | 0.13 | 0.046 | 46 |
| 861 | 3.4 | 0.19 | 0.06 | 120 | 0.19 | 0.068 | 43 | 0.12 | 0.044 | 54 |
| **1181** | 4.5 | 0.19 | 0.06 | 120 | **0.18** | 0.064 | **50** | 0.12 | 0.041 | 60 |
| 1501 | 5.4 | 0.19 | 0.06 | 120 | 0.17 | 0.062 | 56 | 0.12 | 0.041 | 65 |
| 2427 | 8.0 | 0.19 | 0.06 | 120 | 0.16 | 0.058 | 73 | 0.12 | 0.039 | 82 |
| 3062 | 9.5 | 0.19 | 0.06 | 120 | 0.16 | 0.057 | 83 | 0.12 | 0.038 | 92 |
| 4598 | 12.2 | 0.19 | 0.06 | 120 | 0.16 | 0.053 | 110 | 0.11 | 0.036 | 120 |

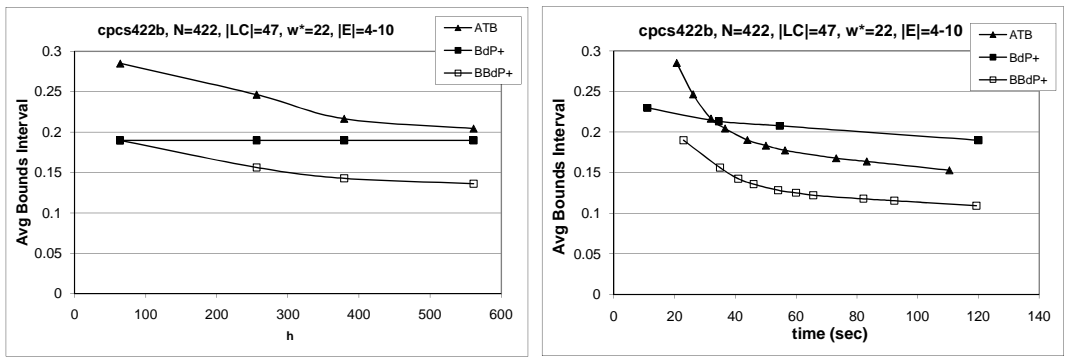

Figure 8: Results for cpcs422b. The table reports the average bounds interval $\overline{I}$, average error $\Delta$, computation time (in seconds), and percent of probability of evidence $P(e)$ covered by the fully-instantiated cutset tuples as a function of $h$. We highlight in bold face the first $ATB$ data point where the average bounds interval is as good or better than $BdP+$. The charts show the convergence of the bounds interval length as a function of $h$ and time.

Table 2 summarizes the characteristics of each network. For each one, the table specifies the number of variables $N$, the induced width $w^*$, the size of loop cutset $|LC|$, the number of loop-cutset tuples $|\mathcal{D}(LC)|$, and the time needed to compute the exact posterior marginals by bucket-tree elimination (exponential in the induced width $w^*$) and by cutset conditioning (exponential in the size of loop-cutset). An 'out' indicates that bucket-tree elimination is unfeasible in terms of memory demands. Note that the characteristics of grid networks only depend on their sizes but not on the percentage of determinism; the characteristics of all coding networks are the same.

For our purposes, we consider $VEC$ as another exact algorithm to compute the exact $P(e)$ in the first and second benchmarks and as a lower bounding technique for the third benchmark. We fix the control parameter $z$ of $MBE$ and the $w$-cutset of $VEC$ so that the algorithms require less than 1.5GB of space.

MUNIN3

| Munin3, N=1044, $w^*$=7, $|LC|$=30, $|E|$=257 | | | | | | | | | | |
|---|---|---|---|---|---|---|---|---|---|---|
| | | BdP+ | | | ATB | | | BBdP+ | | |
| h | %P(e) | $\bar{I}$ | $\Delta$ | $time(sec)$ | $\bar{I}$ | $\Delta$ | $time(sec)$ | $\bar{I}$ | $\Delta$ | $time(sec)$ |
| **196** | 64 | 0.24 | 0.1 | 12 | **0.050** | 0.020 | **8** | 0.048 | 0.020 | 16 |
| 441 | 72 | 0.24 | 0.1 | 12 | 0.030 | 0.011 | 12 | 0.029 | 0.012 | 20 |
| 882 | 78 | 0.24 | 0.1 | 12 | 0.025 | 0.009 | 18 | 0.025 | 0.009 | 26 |
| 1813 | 79 | 0.24 | 0.1 | 12 | 0.020 | 0.007 | 32 | 0.019 | 0.007 | 40 |
| 2695 | 80 | 0.24 | 0.1 | 12 | 0.018 | 0.006 | 46 | 0.017 | 0.007 | 54 |
| 2891 | 81 | 0.24 | 0.1 | 12 | 0.017 | 0.006 | 49 | 0.016 | 0.006 | 57 |
| 3185 | 82 | 0.24 | 0.1 | 12 | 0.014 | 0.005 | 54 | 0.014 | 0.005 | 62 |
| 3577 | 82 | 0.24 | 0.1 | 12 | 0.013 | 0.004 | 68 | 0.012 | 0.004 | 76 |
| 4312 | 83 | 0.24 | 0.1 | 12 | 0.011 | 0.004 | 80 | 0.010 | 0.004 | 88 |

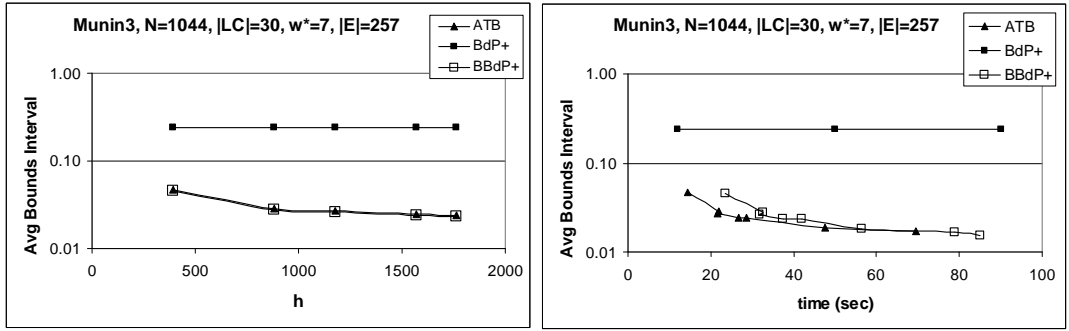

MUNIN4

| Munin4, N=1041, $w^*$=8, $|LC|$=49, $|E|$=235 | | | | | | | | | | |
|---|---|---|---|---|---|---|---|---|---|---|
| | | BdP+ | | | ATB | | | BBdP+ | | |
| h | %P(e) | $\bar{I}$ | $\Delta$ | $time(sec)$ | $\bar{I}$ | $\Delta$ | $time$ | $\bar{I}$ | $\Delta$ | $time(sec)$ |
| 245 | 1 | 0.23 | 0.1 | 15 | 0.39 | 0.16 | 14 | 0.24 | 0.102 | 21 |
| 441 | 7 | 0.23 | 0.1 | 15 | 0.32 | 0.13 | 17 | 0.22 | 0.095 | 24 |
| 1029 | 11 | 0.23 | 0.1 | 15 | 0.28 | 0.12 | 34 | 0.21 | 0.089 | 44 |
| **2058** | 17 | 0.23 | 0.1 | 15 | **0.25** | 0.11 | 54 | 0.19 | 0.082 | 65 |
| 3087 | 20 | 0.23 | 0.1 | 15 | 0.22 | 0.11 | 83 | 0.18 | 0.077 | 91 |
| 5194 | 24 | 0.23 | 0.1 | 15 | 0.21 | 0.09 | 134 | 0.17 | 0.072 | 145 |

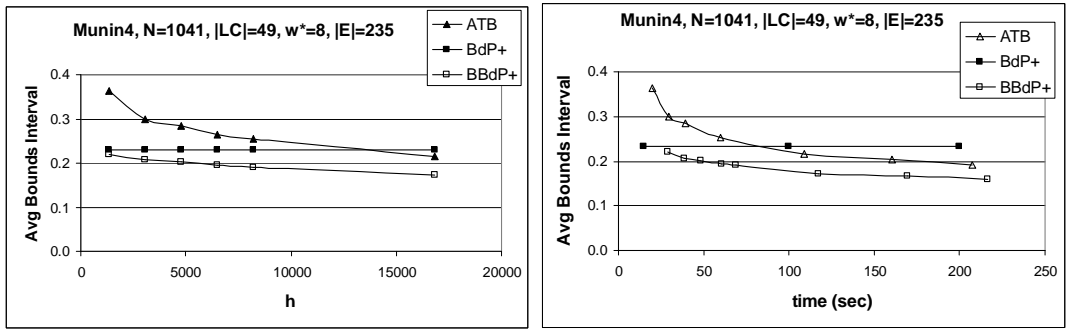

Figure 9: Results for munin3 and munin4. The tables report the average bounds interval $\bar{I}$, average error $\Delta$, computation time (in seconds), and percent of probability of evidence $P(e)$ covered by the fully-instantiated cutset tuples as a function of $h$. We highlight in bold face the first $ATB$ data point where the average bounds interval is as good or better than $BdP+$. The charts show the convergence of the bounds interval length as a function of $h$ and time.

| network | N | w* | \|LC\| | $\|\mathcal{D}(LC)\|$ | \|E\| | Time(BE) | Time(LC) |
|---|---|---|---|---|---|---|---|
| bn2o-15-30-15 | 45 | 22 | 24 | $2^{24}$ | 15 | 14.51 | 17.4 hrs |
| bn2o-15-30-20 | 50 | 25 | 26 | $2^{26}$ | 20 | 174.28 | 93.2 hrs |
| bn2o-15-30-25 | 55 | 24 | 25 | $2^{25}$ | 25 | 66.23 | 75.76 hrs |
| Grids | $16 \times 16$ | 22 | 116 | $2^{116}$ | 1 | 27.59 | $> 2^{93}$ hrs[1] |
| Grids | $20 \times 20$ | 29 | 185 | $2^{185}$ | 1 | out | $> 2^{66}$ hrs[1] |
| Grids | $26 \times 26$ | 40 | 325 | $2^{325}$ | 1 | out | $> 2^{306}$ hrs[1] |
| Grids | $42 \times 42$ | 70 | 863 | $2^{863}$ | 1 | out | $> 2^{844}$ hrs[1] |
| coding | 512 | 54-61 | 59-64 | $2^{59}$-$2^{64}$ | 256 | out | $> 2^{42}$ hrs[1] |

Table 2: Complexity characteristics of the benchmarks from the UAI repository: $N$-number of nodes, $w^*$-induced width, $|LC|$-number of nodes in a loop-cutset, $|\mathcal{D}(LC)|$-loop-cutset state space size, Time(BE) is the exact computation time via bucket elimination, Time(LC) is the exact computation time via loop-cutset conditioning. The results are averaged over the set of network instances of each benchmark.

### 4.3.2 RESULTS

We summarize the results for each benchmark in a tabular format. The tables report the bounds and computation time (in seconds) for each compared algorithm. For $ATB$, we report results by varying the values of the control parameters $(h, k)$. In particular, we consider values of $h$ in the range 4 to 200, and values of $k$ in the set $\{2^{10}, 2^{12}, 2^{14}\}$. By doing so, we analyze the impact of each control parameter on the performance of the algorithm. Grey areas in the tables correspond to $(h, k)$ configurations that cannot be compared due to computation time.

**Two-layer noisy-or networks**. Table 3 shows the results. As expected, the quality of the bounds produced by $ATB$ improves when the values of the control parameters $(h, k)$ increase. We observe that the best bounds are obtained when fixing $h$ to the highest value (i.e., 200) and $k$ to the smallest value (i.e., $2^{10}$). However, the increase in the value of $h$ leads to higher computation times than when increasing the value of $k$. When taking time into account, comparing configurations with similar time (see $(h = 50, k = 2^{10})$ and $(h = 4, k = 2^{14})$, and $(h = 200, k = 2^{10})$ and $(h = 50, k = 2^{12})$, respectively), we observe that the configuration with the highest value of $h$ and the smallest value of $k$ outperforms the other ones.

When compared with $MBE$, there is no clear superior approach. The accuracy of the algorithms depends on whether we look at upper or lower bounds. When considering upper bounds, $ATB$ outperforms $MBE$ for all instances 1b, 2b and 3b. Note that for those instances, $MBE$ computes worse upper bounds than the trivial one (i.e., greater than 1). However, for instances 1a, 2a and 3a, $MBE$ computes tighter upper bounds than $ATB$. For lower bounds, in general $ATB$ outperforms MBE for instances with 20 and 25 evidence variables, while $MBE$ is more accurate for instances having 15 evidence variables. Regarding computation time, $ATB$ is definitely slower than $MBE$.

---

1. Times are extrapolated.

| Inst. | P(e) | h | %P(e) | ATB($h, k = 2^{10}$) LB/UB | Time | ATB($h, k = 2^{12}$) LB/UB | Time | ATB($h, k = 2^{14}$) LB/UB | Time | MBE(z=18) LB/UB | Time |
|---|---|---|---|---|---|---|---|---|---|---|---|
| | | | | bn2o-30-15-150, $|E| = 15$ | | | | | | | |
| 1a | 5.9E-05 | 4 | 0.0004 | 5.7E-10/5.3E-01 | 2 | 5.9E-10/4.8E-01 | 8 | 6.0E-10/4.4E-01 | 38 | | |
| | | 50 | 0.100 | 1.2E-07/1.1E-01 | 32 | 1.2E-07/9.4E-02 | 129 | | | 9.1E-06/4.8E-04 | 2 |
| | | 200 | 0.250 | 3.5E-07/5.7E-02 | 103 | | | | | | |
| 1b | 0.56565 | 4 | 0.007 | 3.1E-04/9.3E-01 | 2 | 4.3E-04/9.3E-01 | 8 | 5.0E-04/9.3E-01 | 38 | | |
| | | 50 | 0.120 | 4.1E-03/8.6E-01 | 31 | 4.8E-03/8.6E-01 | 124 | | | 0.17277/1.42 | 2 |
| | | 200 | 0.460 | 1.5E-02/8.5E-01 | 102 | | | | | | |
| 2a | 4.0E-07 | 4 | 0.003 | 2.0E-11/1.3E-01 | 2 | 2.0E-11/1.1E-01 | 8 | 2.1E-11/8.5E-02 | 38 | | |
| | | 50 | 0.020 | 1.5E-10/1.0E-02 | 29 | 1.5E-10/8.9E-03 | 115 | | | 8.4E-10/2.1E-05 | 2 |
| | | 200 | 0.320 | 1.7E-09/4.0E-03 | 89 | | | | | | |
| 2b | 0.54111 | 4 | 0.008 | 6.9E-03/7.9E-01 | 2 | 8.6E-03/8.0E-01 | 8 | 9.2E-03/8.0E-01 | 38 | | |
| | | 50 | 0.210 | 5.5E-02/7.8E-01 | 29 | 6.1E-02/7.7E-01 | 115 | | | 0.02647/1.8 | 2 |
| | | 200 | 1.110 | 1.1E-01/7.5E-01 | 90 | | | | | | |
| 3a | 1.2E-04 | 4 | 0.216 | 2.9E-07/1.7E-01 | 2 | 2.9E-07/1.5E-01 | 8 | 2.9E-07/1.4E-01 | 38 | | |
| | | 50 | 1.040 | 1.7E-06/4.6E-02 | 26 | 1.7E-06/4.2E-02 | 103 | | | 4.4E-07/1.5E-03 | 2 |
| | | 200 | 3.580 | 5.3E-06/2.7E-02 | 74 | | | | | | |
| 3b | 0.18869 | 4 | 0.076 | 1.1E-03/7.7E-01 | 2 | 1.1E-03/7.7E-01 | 8 | 1.2E-03/7.7E-01 | 38 | | |
| | | 50 | 0.470 | 6.8E-03/6.3E-01 | 25 | 7.1E-03/6.3E-01 | 95 | | | 0.03089/0.81 | 2 |
| | | 200 | 1.440 | 2.1E-02/5.4E-01 | 69 | | | | | | |
| | | | | bn2o-30-20-200, $|E| = 20$ | | | | | | | |
| 1a | 1.4E-07 | 4 | 0.004 | 5.4E-12/1.6E-02 | 3 | 5.4E-12/1.5E-02 | 16 | 5.4E-12/1.4E-02 | 67 | | |
| | | 50 | 0.050 | 9.1E-11/1.8E-03 | 62 | 9.1E-11/1.6E-03 | 264 | | | 2.4E-15/3.3E-04 | 3 |
| | | 200 | 1.880 | 2.8E-09/5.7E-04 | 195 | | | | | | |
| 1b | 0.15654 | 4 | 0.012 | 1.0E-04/7.3E-01 | 3 | 1.1E-04/7.3E-01 | 16 | 1.1E-04/7.3E-01 | 68 | | |
| | | 50 | 0.140 | 3.3E-03/6.7E-01 | 64 | 3.6E-03/6.7E-01 | 279 | | | 9.8E-04/1.9 | 3 |
| | | 200 | 0.430 | 1.1E-02/5.9E-01 | 218 | | | | | | |
| 2a | 2.2E-07 | 4 | 0.013 | 3.8E-11/1.6E-02 | 3 | 3.8E-11/1.6E-02 | 16 | 3.8E-11/1.5E-02 | 70 | | |
| | | 50 | 0.410 | 1.4E-09/3.3E-03 | 52 | 1.3E-09/3.1E-03 | 211 | | | 4.4E-15/8.0E-05 | 3 |
| | | 200 | 1.410 | 4.5E-09/2.4E-03 | 169 | | | | | | |
| 2b | 0.27695 | 4 | 0.020 | 6.4E-03/8.3E-01 | 3 | 7.3E-03/8.3E-01 | 16 | 8.0E-03/8.3E-01 | 68 | | |
| | | 50 | 0.430 | 3.0E-02/7.7E-01 | 51 | 3.3E-02/7.7E-01 | 197 | | | 2.3E-05/2.9 | 3 |
| | | 200 | 1.620 | 5.9E-02/6.9E-01 | 145 | | | | | | |
| 3a | 2.4E-09 | 4 | 0.002 | 8.3E-14/1.8E-03 | 3 | 8.3E-14/1.8E-03 | 16 | 8.3E-14/1.8E-03 | 68 | | |
| | | 50 | 0.060 | 2.2E-12/1.1E-04 | 58 | 2.2E-12/1.1E-04 | 236 | | | 5.2E-13/1.7E-06 | 3 |
| | | 200 | 0.090 | 3.6E-12/3.3E-05 | 198 | | | | | | |
| 3b | 0.48039 | 4 | 0.0002 | 4.5E-05/9.7E-01 | 3 | 5.1E-05/9.7E-01 | 16 | 6.2E-05/9.7E-01 | 68 | | |
| | | 50 | 0.110 | 5.4E-02/9.3E-01 | 64 | 5.9E-02/9.3E-01 | 277 | | | 5.3E-03/1.9 | 3 |
| | | 200 | 0.660 | 1.1E-01/8.8E-01 | 194 | | | | | | |
| | | | | bn2o-30-25-250, $|E| = 25$ | | | | | | | |
| 1a | 2.9E-09 | 4 | 0.0004 | 1.3E-14/6.6E-02 | 6 | 1.3E-14/6.5E-02 | 22 | 1.3E-14/4.8E-02 | 99 | | |
| | | 50 | 0.01 | 3.7E-13/3.3E-03 | 119 | 3.7E-13/2.8E-03 | 429 | | | 1.7E-16/3.1E-06 | 4 |
| | | 200 | 0.06 | 2.0E-12/1.1E-03 | 396 | | | | | | |
| 1b | 0.15183 | 4 | 0.016 | 4.3E-04/8.1E-01 | 6 | 5.7E-04/8.1E-01 | 22 | 6.2E-04/8.1E-01 | 99 | | |
| | | 50 | 0.22 | 4.6E-03/7.2E-01 | 120 | 6.7E-03/7.2E-01 | 437 | | | 1.4E-03/1.4 | 4 |
| | | 200 | 1.07 | 1.3E-02/6.5E-01 | 381 | | | | | | |
| 2a | 2.4E-07 | 4 | 0.0004 | 1.8E-12/1.9E-01 | 6 | 1.8E-12/1.9E-01 | 22 | 1.8E-12/1.7E-01 | 99 | | |
| | | 50 | 0.0012 | 5.7E-12/4.5E-02 | 112 | 5.7E-12/3.9E-02 | 398 | | | 1.8E-12/1.2E-05 | 4 |
| | | 200 | 0.07 | 1.8E-10/2.2E-02 | 402 | | | | | | |
| 2b | 0.30895 | 4 | 0.018 | 5.3E-04/7.6E-01 | 6 | 5.9E-04/7.6E-01 | 22 | 6.3E-04/7.6E-01 | 99 | | |
| | | 50 | 0.19 | 5.4E-03/7.4E-01 | 107 | 6.1E-03/7.4E-01 | 374 | | | 7.2E-03/1.7 | 4 |
| | | 200 | 0.65 | 1.4E-02/7.1E-01 | 367 | | | | | | |
| 3a | 2.7E-10 | 4 | 0.0001 | 1.7E-16/1.1E-01 | 6 | 1.7E-16/1.1E-01 | 22 | 1.7E-16/8.1E-02 | 99 | | |
| | | 50 | 0.01 | 4.3E-14/1.9E-02 | 119 | 4.3E-14/1.6E-02 | 427 | | | 1.3E-15/4.9E-07 | 4 |
| | | 200 | 0.20 | 5.5E-13/7.1E-03 | 409 | | | | | | |
| 3b | 0.46801 | 4 | 0.0065 | 1.0E-03/7.9E-01 | 6 | 1.2E-03/7.9E-01 | 22 | 1.3E-03/7.9E-01 | 98 | | |
| | | 50 | 0.45 | 4.2E-02/7.7E-01 | 106 | 4.8E-02/7.7E-01 | 352 | | | 3.5E-03/2.7 | 4 |
| | | 200 | 1.52 | 8.5E-02/7.5E-01 | 337 | | | | | | |

Table 3: Results for bn2o networks. The table shows the LB and UB computed by *ATB* varying the number of cutset tuples $h$ and the maximum domain $k$ of the Markov boundary.

| | | | | Grids2, $|E| = 1$ | | | | | | | |
|---|---|---|---|---|---|---|---|---|---|---|---|
| $(M,D)$ | P(e) | h | %P(e) | ATB($k=2^{10}$,h) | | ATB($k=2^{12}$,h) | | ATB($k=2^{14}$,h) | | MBE | |
| | | | | LB/UB | Time | LB/UB | Time | LB/UB | Time | LB / UB | Time |
| | | 4 | 1.57E-14 | 0.3127/0.8286 | 1 | 0.3127/0.8286 | 1 | 0.3127/0.8286 | 1 | | |
| (16, 50) | 0.6172 | 100 | 3.50E-11 | 0.3127/0.8286 | 57 | 0.3127/0.8286 | 57 | | | 0/5.13 | 16 |
| | | 200 | 4.22E-11 | 0.3127/0.8286 | 111 | | | | | | |
| | | 4 | 1.07E-24 | 0.1765/0.4939 | 5 | 0.1765/0.4939 | 5 | 0.1765/0.4939 | 5 | | |
| (20, 50) | 0.4441 | 100 | 1.57E-21 | 0.1765/0.4939 | 208 | 0.1765/0.4939 | 203 | | | 0/12411 | 39 |
| | | 200 | 1.13E-20 | 0.1765/0.4939 | 412 | | | | | | |
| | | 4 | 1.25E-09 | 0.2106/0.7454 | 3 | 0.2106/0.7454 | 3 | 0.2106/0.7454 | 3 | | |
| (20, 75) | 0.4843 | 100 | 2.40E-09 | 0.2106/0.7454 | 81 | 0.2106/0.7454 | 80 | | | 0/1E+05 | 39 |
| | | 200 | 2.89E-09 | 0.2106/0.7454 | 156 | | | | | | |
| | | 4 | 3.88E-19 | 0.0506/0.9338 | 6 | 0.0506/0.9338 | 6 | 0.0506/0.9338 | 6 | | |
| (26, 75) | 0.6579 | 100 | 7.32E-19 | 0.0506/0.9338 | 268 | 0.0506/0.9338 | 270 | | | 0/1E+10 | 84 |
| | | 200 | 1.55E-18 | 0.0506/0.9338 | 534 | | | | | | |
| | | 4 | 3.47E-08 | 0.1858/0.8943 | 2 | 0.1858/0.8943 | 2 | 0.1858/0.8943 | 2 | | |
| (26, 90) | 0.8206 | 100 | 3.41E-06 | 0.1858/0.8943 | 85 | 0.1858/0.8943 | 84 | | | 0/1E+10 | 87 |
| | | 200 | 8.38E-06 | 0.1858/0.8943 | 164 | | | | | | |
| | | 4 | 8.65E-29 | 0.0048/0.9175 | 10 | 0.0048/0.9175 | 10 | 0.0048/0.9175 | 10 | | |
| (42, 90) | 0.4933 | 100 | 2.32E-25 | 0.0048/0.9175 | 436 | 0.0048/0.9175 | 439 | | | 0/1E+10 | 110 |
| | | 200 | 3.48E-25 | 0.0048/0.9175 | 866 | | | | | | |

Table 4: Results on grid networks. The table shows the LB and UB computed by $ATB$ varying the number of cutset tuples $h$ and the maximum length $k$ of the conditional probability tables over the Markov boundary.

**Grid networks**. Table 4 reports the results. The first thing to observe is that $MBE$ computes completely uninformative bounds. In this case, the anytime behavior of $ATB$ is not effective either. The increase of the value of its control parameters $(h, k)$ does not affect its accuracy. Since the Markov boundary in grid networks is relatively small, the smallest tested value of $k$ is higher than its Markov boundary size which explains the independence on $k$. Another reason for its ineffectiveness may be the high percentage of determinism in these networks. It is known that sampling methods are inefficient in the presence of determinism. As a consequence, the percentage of probability mass accumulated in the $h$ sampled tuples is not significant, which cancels the benefits of computing exact probability of evidence for that subset of tuples. Therefore, in such cases a more sophisticated sampling scheme should be used, for example (Gogate & Dechter, 2007). Consequently, for these deterministic grids, $ATB$'s performance is controlled totally by its bound propagation plugged-in algorithm.

**Coding networks**. Table 5 shows the results. We do not report the percentage of $P(e)$ covered by the fully-instantiated cutset tuples because the exact $P(e)$ is not available. We set the time limit of $VEC$ to 1900 seconds (i.e., the maximum computation time required by running $ATB$ in these instances). We only report the results for $k=2^{10}$ and $k=2^{14}$ because the increase in the value of $k$ was not effective and did not result in increased accuracy. In this case, the accuracy of $ATB$ increases as the value of $h$ increases. In comparing $ATB$ with the other algorithms we have to distinguish between lower and upper bounds. Regarding lower bounds, $ATB$ clearly outperforms $MBE$ and $VEC$ in all instances. Indeed, the lower bound computed by $MBE$ and $VEC$ is very loose. Regarding upper bounds, $ATB(h = 150, k = 2^{10})$ outperforms $MBE$ in three instances (i.e., $BN\_128$, $BN\_129$ and $BN\_131$). When taking time into account $ATB$ only outperforms $MBE$ in instance $BN\_129$.

| | | coding, $|E| = 256$ | | | | | | | |
|---|---|---|---|---|---|---|---|---|---|
| Inst. | h | ATB($k=2^{10}$,h) | | ATB($k=2^{14}$,h) | | MBE(z=22) | | VEC | |
| | | LB/UB | Time | LB/UB | Time | LB/UB | Time | LB | Time |
| | 4 | 1.9E-76/1.5E-41 | 50 | 1.9E-76/1.52E-41 | 3494 | | | | |
| BN_126 | 50 | 1.9E-69/2.5E-42 | 632 | | | 1.4E-139/1.5E-044 | 143 | 9.2E-102 | 1900 |
| | 150 | 1.9E-58/1.3E-42 | 1442 | | | | | | |
| | 4 | 5.3E-60/2.3E-43 | 55 | 5.3E-60/2.3E-43 | 399 | | | | |
| BN_127 | 50 | 1.3E-58/2.3E-44 | 426 | | | 1.6E-134/1.0E-045 | 164 | 5.3E-115 | 1900 |
| | 150 | 1.6E-58/1.9E-44 | 946 | | | | | | |
| | 4 | 7.2E-54/1.6E-42 | 85 | 7.2E-54/1.6E-42 | 582 | | | | |
| BN_128 | 50 | 4.9E-48/7.2E-43 | 637 | | | 1.2E-144/5.1E-043 | 124 | 1.9E-112 | 1900 |
| | 150 | 4.9E-48/1.4E-43 | 1225 | | | | | | |
| | 4 | 1.4E-72/8.2E-45 | 50 | 1.5E-72/8.2E-45 | 362 | | | | |
| BN_129 | 50 | 1.5E-64/2.1E-45 | 585 | | | 2.8E-139/4.8E-043 | 144 | 1.5E-115 | 1900 |
| | 150 | 8.5E-64/5.4E-46 | 1400 | | | | | | |
| | 4 | 4.7E-65/2.9E-44 | 47 | 4.7E-65/2.9E-44 | 324 | | | | |
| BN_130 | 50 | 6.3E-63/2.9E-45 | 619 | | | 1.1E-132/1.9E-045 | 112 | 1.3E-96 | 1900 |
| | 150 | 3.7E-58/2.3E-45 | 1299 | | | | | | |
| | 4 | 1.9E-60/1.3E-44 | 52 | 1.9E-60/1.3E-44 | 367 | | | | |
| BN_131 | 50 | 2.3E-54/3.7E-45 | 484 | | | 2.3E-141/3.2E-045 | 119 | 3.2E-102 | 1900 |
| | 150 | 2.3E-51/1.1E-45 | 1276 | | | | | | |
| | 4 | 2.3E-79/6.3E-44 | 50 | 2.3E-79/6.3E-44 | 363 | | | | |
| BN_132 | 50 | 3.6E-67/1.0E-44 | 689 | | | 2.8E-134/2.3E-048 | 109 | 8.9E-111 | 1900 |
| | 150 | 1.5E-66/8.1E-45 | 1627 | | | | | | |
| | 4 | 1.6E-56/2.7E-42 | 53 | 1.6E-56/2.7E-42 | 398 | | | | |
| BN_133 | 50 | 1.1E-54/2.4E-43 | 671 | | | 1.8E-136/4.1E-045 | 147 | 1.9E-109 | 1900 |
| | 150 | 2.3E-54/9.5E-44 | 1846 | | | | | | |
| | 4 | 8.9E-63/1.8E-43 | 47 | 8.9E-63/1.8E-43 | 355 | | | | |
| BN_134 | 50 | 1.2E-62/8.6E-45 | 606 | | | 1.9E-148/3.9E-045 | 163 | 4.2E-111 | 1900 |
| | 150 | 6.1E-57/4.8E-45 | 1412 | | | | | | |

Table 5: Results on coding networks. The table shows the LB and UB computed by $ATB$ varying the number of cutset tuples $h$ and the maximum length $k$ of the conditional probability tables over the Markov boundary.

**Summary of empirical evaluation.** We demonstrated that $ATB$'s bounds converge as $h$, the number of cutset tuples computed exactly, increases. The speed of convergence varied among benchmarks. The convergence was faster when the active cutset tuples accounted for a large percentage of the probability mass of $P(C|e)$, as shown for the case of $cpcs54$, $cpcs179$, and $cpcs360$ networks. Comparing with a variant of bound propagation called $BdP+$, $ATB$ was more accurate if given sufficient time and even when given the same time bound, it computed more accurate bounds on many benchmarks.

We showed that $ATB$'s bounds on the posterior marginals can be further improved when used as initial bounds in $BdP+$. We call this hybrid of $ATB$ followed by $BdP+$ the $BBdP+$ algorithm. Our experiments demonstrated the added power of $BBdP+$ in exploiting the time-accuracy trade-off.

We also compared the power of $ATB$ to bound the probability of evidence against the mini-bucket elimination ($MBE$). We showed that neither algorithm was dominating on all benchmarks. Given the same amount of time, $ATB$ computed more accurate bounds than $MBE$ on some instances of *bn2o* and *coding networks*. $ATB$ outperformed $MBE$ on all instances of the *grid networks* on which $MBE$ only computed bounds of 0 and 1. On this benchmark, however, $ATB$ converged very slowly. We believe in part this is due to the grid's large loop-cutset sizes.

We compared $ATB$'s ability to compute the lower bound on $P(e)$ to $VEC$ on coding networks. $VEC$ obtains the bound by computing a partial sum in the cutset-conditioning formula (see Eq. 2). By comparing the lower bounds generated by $ATB$ and $VEC$, we can gain insight into the trade-off between enumerating more cutset tuples and bounding the uninstantiated tuples. Since $ATB$'s lower bound was consistently tighter, we conclude that bounding the uninstantiated tuples is cost-effective.

## 5. Related Work

There are three early approaches which use the same principle as $ATB$: Poole's algorithm (1996), *bounded conditioning* ($BC$) (Horvitz et al., 1989) which we have already described, and *bounded recursive decomposition* (Monti & Cooper, 1996). In all these cases the computation of the bounds is composed of an exact inference over a subset of the tuples and a bounding scheme over the total probabilities over the rest of the tuples.

Similar to $ATB$, Poole's scheme is based on a partial exploration of a search tree. However, his search tree corresponds to the state space over all the variables of the whole network and hence, it is exponential in the total number of variables. In contrast, the tree structure used by our approach corresponds to the state space of the loop-cutset variables; therefore, it is exponential in the loop-cutset size only. In addition, Poole *updates* the bounding function when a tuple with probability 0 (i.e., a conflict) is discovered.

As discussed in Section 2.2, $BC$ is also based on the cutset conditioning principle, but there are two main differences relative to $ATB$: (i) the probability mass of the missing tuples is bounded via prior probabilities, and consequently (ii) as we proved, the upper bound expression is looser.

*Bounded recursive decomposition* uses Stochastic simulation (Pearl, 1988) to generate highly probable instantiations of the variables, which is similar to $ATB$, and bounds the missing elements with 0 and prior values. Therefore this approach resembles Poole's algorithm and bounded conditioning. Unlike $ATB$, bounded recursive decomposition requires instantiation of all the variables in the network and relies on priors to guide the simulation. In contrast, our algorithm uses Gibbs sampling on a cutset only which is likely to be more accurate at selecting high probability tuples in presence of evidence. $ATB$ subsumes all three algorithms offering a unifying approach to bounding posteriors with anytime properties, able to improve its bounds by investing more time and exploring more cutset tuples.

There are a number of alternative approaches for computing bounds on the marginals. Poole (1998) proposed *context-specific* bounds obtained from simplifying the conditional probability tables. The method performs a variant of bucket elimination where intermediate tables are collapsed by grouping some probability values together. However, since the method was validated only on a small car diagnosis network with 10 variables, it is hard to draw any conclusions. Larkin (2003) also obtains bounds by simplifying intermediate probability tables in the variable elimination order. He solves an optimization problem to find a table decomposition that minimizes the error. Kearns and Saul (1999, 1998) proposed a specialized *large deviation bounds* approach for layered networks, while Mannino and Mookerjee (2002) suggested an elaborate bounding scheme with nonlinear objective functions. Jaakkola and Jordan (1999) proposed a variational method for computing lower and upper bounds on posterior marginals in Noisy-Or networks and evaluated its performance in the case of diagnostic QMR-DT network. More recent approaches (Tatikonda, 2003; Taga & Mase, 2006; Ihler, 2007; Mooij & Kappen, 2008) aim to bound the error of *belief propagation* marginals. The first two approaches are exponential in the size of the Markov boundary. The third approach is linear in the size of the network, but is formulated for pairwise interactions only. Finally, the fourth algorithm is exponential in the number of domain values. Recently, Mooij and Kappen (2008) proposed the *box propagation* algorithm that propagates local bounds (convex sets of probability distributions) over a subtree of the factor graph representing the problem, rooted in the variable of interest.

It is important to note that our approach offers an anytime framework for computing bounds where any of the above bounding algorithms can be used as a subroutine to bound joint probabilities for partially-instantiated tuples within $ATB$ and therefore may improve the performance of any bounding scheme.

Regarding algorithms that bound the probability of evidence, we already mentioned the mini-bucket schemes and compared against it in Section 4.3. Another recent approach is the *tree-reweighted belief propagation* ($TRW$-$BP$) (Wainwright, Jaakkola, & Willsky, 2005). $TRW$-$BP$ is a class of message-passing algorithms that compute an upper bound of $P(e)$ as a convex combination of tree-structured distributions. In a recent paper, Rollon and Dechter (2010) compare $TRW$-$BP$, box propagation (adapted for computing the probability of evidence using the chain rule), $MBE$ and $ATB$-$ABdP+$. Their empirical evaluation shows the relative strength of each scheme on the different benchmarks (Rollon & Dechter, 2010). In another recent work Wexler and Meek (2008) have proposed MAS, a bounding algorithm for computing the probability of evidence. Shekhar (2009) describes the adjustments required to produce bounds using MAS for Bayesian networks, where the potentials are less than 1. In a forthcoming paper, Wexler and Meek (2010) improve their MAS scheme to obtain tighter bounds and describe how to obtain bounds for Bayesian networks for $P(e)$ as well as for other inferential problems such as the maximal *a posteriori* and most probable explanation problems. The comparison with this approach is left as future work.

## 6. Summary and Conclusions

The paper explores a general theme of approximation and bounding algorithms for likelihood computation, a task that is known to be hard. While a few methods based on one or two principles emerge, it is clear that pooling together a variety of ideas into a single framework can yield a significant improvement. The current paper provides such a framework. It utilizes the principle of cutset conditioning harnessing the varied strengths of different methods. The framework is inherently anytime, an important characteristic for approximation schemes.

Cutset conditioning is a universal principle. It allows decomposing a problem into a collection of more tractable ones. Some of these subproblems can be solved exactly while others can be approximated. The scheme can be controlled by several parameters. In $w$- cutset we condition on a subset of variables until their treewidth is bounded by $w$. Each subproblem can then be solved exactly in time and space exponential in $w$. If the number of subproblems is too large, we can use another parameter, $h$, to control the number of subproblems solved exactly. The rest of the subproblems are solved using an off-the-shelf bounding scheme.

We developed an expression that incorporates all these aspects using the parameters: $w$ - the induced-width of the cutset, $h$ - the number of cutset conditioning subproblems to be solved exactly (e.g., by bucket elimination), and $\mathcal{A}$ - the approximation algorithm that bounds each of the bounded subproblems. We showed that the number of subproblems that are approximated is polynomial in $h$.

In our empirical evaluation of the general framework, called $ATB$, we used the loop-cutset scheme ($w = 1$) and chose as a bounding algorithm a variant of bound propagation (Leisink & Kappen, 2003), yielding the integrated scheme which we call $ABdP+$. We experimented with several benchmarks for the computing posterior marginals and the probability of evidence, and compared against relevant state of the art algorithms.

Our results demonstrate the value of our $ATB$ framework across all the benchmarks we have tried. As expected, its anytime aspect is visible showing improved accuracy as a function of time. More significantly, even when provided with equal time and space resources, $ATB$ showed remarkable superiority when compared with our variant of bound propagation and with the mini-bucket elimination algorithm ($MBE$) (Dechter & Rish, 2003). The latter was recently investigated further by Rollon and Dechter (2010).

Overall, we can conclude that $ATB$ is a competitive algorithm for both bounding posterior marginals and probability of evidence. Generally, we can expect $ATB$ to perform well in networks whose cutset $C$ is small relative to the total number of variables and whose distribution $P(C|e)$ has a small number of high probability tuples.

The possibilities for future work are many. We can explore additional trade offs such as increasing $w$ and therefore decreasing $h$ and improving the selection of the $h$ tuples. We have looked at only one possible instantiation of the plug-in algorithm $\mathcal{A}$. Other approximation algorithms can be tried which may offer different time/accuracy trade-offs. In particular, we plan to investigate the effectiveness of $ATB$ using $MBE$ as plug-in algorithm.

## Acknowledgments

This work was supported in part by the NSF under award numbers IIS-0331707, IIS-0412854 and IIS-0713118 and by the NIH grant R01-HG004175-02.

Emma Rollon's work was done while a postdoctoral student at the Bren School of Information and Computer Sciences, University of California, Irvine.

The work here was presented in part in (Bidyuk & Dechter, 2006a, 2006b).

## Appendix A. Analysis of Bounded Conditioning

THEOREM **2.1** The interval between lower and upper bounds computed by bounded conditioning is lower bounded by the probability mass of prior distribution $P(C)$ of the unexplored cutset tuples: $\forall h, P_{BC}^U(x|e) - P_{BC}^L(x|e) \geq \sum_{i=h+1}^{M} P(c^i)$.

Proof.

$$
\begin{aligned}
P_{BC}^{U}(x|e) - P_{BC}^{L}(x|e) \quad &= \quad \frac{\sum_{i=h+1}^{M} P(c^i)(\sum_{i=1}^{h} P(c^i, e) + \sum_{i=h+1}^{M} P(c^i))}{\sum_{i=1}^{h} P(c^i, e)} \\
&+ \quad \frac{\sum_{i=1}^{h} P(x, c^i, e)}{\sum_{i=1}^{h} P(c^i, e)} - \frac{\sum_{i=1}^{h} P(x, c^i, e)}{\sum_{i=1}^{h} P(c^i, e) + \sum_{i=h+1}^{M} P(c^i)} \\
&\geq \quad \frac{\sum_{i=h+1}^{M} P(c^i)(\sum_{i=1}^{h} P(c^i, e) + \sum_{i=h+1}^{M} P(c^i))}{\sum_{i=1}^{h} P(c^i, e)} \\
&= \quad \sum_{i=h+1}^{M} P(c^i) + \frac{(\sum_{i=h+1}^{M} P(c^i))^2}{\sum_{i=1}^{h} P(c^i, e)} \geq \sum_{i=h+1}^{M} P(c^i)
\end{aligned}
$$

□

## Appendix B. Bounding Posteriors of Cutset Nodes

So far, we only considered computation of posterior marginals for variable $X \in \mathcal{X} \backslash (C \cup E)$. Now we focus on computing bounds for a cutset node $C_k \in C$. Let $c'_k \in \mathcal{D}(C)$ be some value in domain of $C_k$. Then, we can compute exact posterior marginal $P(c_k|e)$ using Bayes formula:

$$
P(c'_k|e) = \frac{P(c'_k, e)}{P(e)} = \frac{\sum_{i=1}^{M} \delta(c'_k, c^i) P(c^i, e)}{\sum_{i=1}^{M} P(c^i, e)} \tag{23}
$$

where $\delta(c'_k, c^i)$ is a Dirac delta-function so that $\delta(c'_k, c^i) = 1$ iff $c_k^i = c'_k$ and $\delta(c'_k, c^i) = 0$ otherwise. To simplify notation, let $Z = C \backslash Z$. Let $M_k$ denote the number of tuples in state-space of $Z$. Then we can re-write the numerator as:

$$
\sum_{i=1}^{M} \delta(c'_k, c^i) P(c^i, e) = \sum_{i=1}^{M_k} P(c'_k, z^i, e)
$$

and the denominator can be decomposed as:

$$
\sum_{i=1}^{M} P(c^i, e) = \sum_{c_k \in \mathcal{D}(C_k)} \sum_{i=1}^{M_k} P(c'_k, z^i, e)
$$

Then, we can re-write the expression for $P(c'_k|e)$ as follows:

$$
P(c'_k|e) = \frac{\sum_{i=1}^{M_k} P(c'_k, z^i, e)}{\sum_{c_k \in \mathcal{D}(C_k)} \sum_{i=1}^{M_k} P(c_k, z^i, e)} \tag{24}
$$

Let $h_{c_k}$ be the number of full cutset tuples where $c_k^i = c_k$. Then, we can decompose the numerator in Eq. (24) as follows:

$$
\sum_{i=1}^{M_k} P(c'_k, z^i, e) = \sum_{i=1}^{h_{c'_k}} P(c'_k, z^i, e) + \sum_{i=h_{c'_k}+1}^{M_k} P(c'_k, z^i, e)
$$

Similarly, we can decompose the sums in the denominator:

$$\sum_{c_k \in \mathcal{D}(C_k)} \sum_{i=1}^{M_k} P(c_k, z^i, e) = \sum_{c_k \in \mathcal{D}(C_k)} \sum_{i=1}^{h_{c_k}} P(c_k, z^i, e) + \sum_{c_k \in \mathcal{D}(C_k)} \sum_{i=h_{c_k}+1}^{M_k} P(c_k, z^i, e)$$

After decomposition, the Eq. (24) takes on the form:

$$P(c'_k|e) = \frac{\sum_{i=1}^{h_{c'_k}} P(c'_k, z^i, e) + \sum_{i=h_{c'_k}+1}^{M_k} P(c'_k, z^i, e)}{\sum_{c_k \in \mathcal{D}(C_k)} \sum_{i=1}^{h_{c_k}} P(c_k, z^i, e) + \sum_{c_k \in \mathcal{D}(C_k)} \sum_{i=h_{c_k}+1}^{M_k} P(c_k, z^i, e)} \quad (25)$$

Now, for conciseness, we can group together all fully instantiated tuples in the denominator:

$$\sum_{c_k \in \mathcal{D}(C_k)} \sum_{i=1}^{h_{c_k}} P(c_k, z^i, e) = \sum_{i=1}^{h} P(c^i, e)$$

Then, Eq. (25) transforms into:

$$P(c'_k|e) = \frac{\sum_{i=1}^{h_{c'_k}} P(c'_k, z^i, e) + \sum_{i=h_{c'_k}+1}^{M_k} P(c'_k, z^i, e)}{\sum_{i=1}^{h} P(c^i, e) + \sum_{i=h_{c_k}+1}^{M_k} \sum_{c_k \in \mathcal{D}(C_k)} P(c_k, z^i, e)} \quad (26)$$

Now, we can replace each sum $\sum_{i=h_{c'_k}+1}^{M_k}$ over unexplored cutset tuples with a sum over the partially-instantiated cutset tuples. Denoting as $M'_{c_k} = M_k - h_{c_k} + 1$ the number of partially instantiated cutset tuples for $C_k = c_k$, we obtain:

$$P(c'_k|e) = \frac{\sum_{i=1}^{h_{c'_k}} P(c'_k, z^i, e) + \sum_{j=1}^{M'_{c'_k}} P(c'_k, z^j_{1:q_j}, e)}{\sum_{i=1}^{h} P(c^i, e) + \sum_{j=1}^{M'_{c_k}} \sum_{c_k \in \mathcal{D}(C_k)} P(c_k, z^j_{1:q_j}, e)} \quad (27)$$

In order to obtain lower and upper bounds formulation, we will separate the sum of joint probabilities $P(c'_k, z^j_{1:q_j}, e)$ where $C_k = c'_k$ from the rest:

$$P(c'_k|e) = \frac{\sum_{i=1}^{h_{c'_k}} P(c'_k, z^i, e) + \sum_{j=1}^{M'_{c'_k}} P(c'_k, z^j_{1:q_j}, e)}{\sum_{i=1}^{h} P(c^i, e) + \sum_{j=1}^{M'_{c'_k}} P(c'_k, z^j_{1:q_j}, e) + \sum_{j=1}^{M'_{c_k}} \sum_{c_k \neq c'_k} P(c_k, z^j_{1:q_j}, e)} \quad (28)$$

In the expression above, probabilities $P(c_k, z^i, e)$ and $P(c^i, e)$ are computed exactly since they correspond to full cutset instantiations. Probabilities $P(c_k, z^i_{1:q_i}, e)$, however, will be bounded since only partial cutset is observed. Observing that both the numerator and denominator have component $P(c'_k, z^i_{1:q_i}, e)$ and replacing it with an upper bound $P^U(c'_k, z^i_{1:q_i}, e)$ in both the numerator and denominator, we will obtain an upper bound on $P(c'_k|e)$ due to Lemma 3.2:

$$P(c'_k|e) \leq \frac{\sum_{i=1}^{h_{c'_k}} P(c'_k, z^i, e) + \sum_{j=1}^{M'_{c'_k}} P^U_{\mathcal{A}}(c'_k, z^j_{1:q_j}, e)}{\sum_{i=1}^{h} P(c^i, e) + \sum_{j=1}^{M'_{c'_k}} P^U_{\mathcal{A}}(c'_k, z^j_{1:q_j}, e) + \sum_{j=1}^{M'_{c_k}} \sum_{c_k \neq c'_k} P(c_k, z^j_{1:q_j}, e)} \quad (29)$$

Finally, replacing $P(c_k, z_{1:q_j}^j, e)$, $c_k \neq c_k'$, with a lower bound (also increasing fraction value), we obtain:

$$P(c_k'|e) \leq \frac{\sum_{i=1}^{h_{c_k'}} P(c_k', z^i, e) + \sum_{j=1}^{M_{c_k}'} P_{\mathcal{A}}^U(c_k', z_{1:q_j}^j, e)}{\sum_{i=1}^{h} P(c^i, e) + \sum_{j=1}^{M_{c_k}'} P_{\mathcal{A}}^U(c_k', z_{1:q_j}^j, e) + \sum_{j=1}^{M_{c_k}'} \sum_{c_k \neq c_k'} P_{\mathcal{A}}^L(c_k, z_{1:q_j}^j, e)} = P_c^U \quad (30)$$

The lower bound derivation is similar. We start with Eq. (28) and replace $P(c_k', z_{1:q_i}^i, e)$ in the numerator and denominator with a lower bound. Lemma 3.2 guarantees that the resulting fraction will be a lower bound on $P(c_k'|e)$:

$$P(c_k'|e) \geq \frac{\sum_{i=1}^{h_{c_k'}} P(c_k', z^i, e) + \sum_{j=1}^{M_{c_k}'} P_{\mathcal{A}}^L(c_k', z_{1:q_j}^j, e)}{\sum_{i=1}^{h} P(c^i, e) + \sum_{j=1}^{M_{c_k}'} P_{\mathcal{A}}^L(c_k', z_{1:q_j}^j, e) + \sum_{j=1}^{M_{c_k}'} \sum_{c_k \neq c_k'} P(c_k, z_{1:q_j}^j, e)} \quad (31)$$

Finally, grouping $P_{\mathcal{A}}^L(c_k', z_{1:q_j}^j, e)$ and $\sum_{c_k \neq c_k'} P(c_k, z_{1:q_j}^j, e)$ under one sum and replacing $P_{\mathcal{A}}^L(c_k', z_{1:q_j}^j, e) + \sum_{c_k \neq c_k'} P(c_k, z_{1:q_j}^j, e)$ with an upper bound, we obtain the lower bound $P_c^L$:

$$P(c_k'|e) \geq \frac{\sum_{i=1}^{h_{c_k'}} P(c_k', z^i, e) + \sum_{j=1}^{M_{c_k}'} P_{\mathcal{A}}^L(c_k', z_{1:q_j}^j, e)}{\sum_{i=1}^{h} P(c^i, e) + \sum_{j=1}^{M_{c_k}'} UB[P_{\mathcal{A}}^L(c_k', z_{1:q_j}^j, e) + \sum_{c_k \neq c_k'} P(c_k, z_{1:q_j}^j, e)]} = P_c^L \quad (32)$$

where

$$UB[P_{\mathcal{A}}^L(c_k', z_{1:q_j}^j, e) + \sum_{c_k \neq c_k'} P(c_k, z_{1:q_j}^j, e)] = \min \left\{ \begin{array}{l} P_{\mathcal{A}}^L(c_k', z_{1:q_j}^j, e) + \sum_{c_k \neq c_k'} P_{\mathcal{A}}^U(c_k, z_{1:q_j}^j, e) \\ P_{\mathcal{A}}^U(z_{1:q_j}^j, e) \end{array} \right.$$

The lower bound $P_c^L$ is a cutset equivalent of the lower bound $P^L$ obtained in Eq. (15).

With respect to computing bounds on $P(c_k', z_{1:q}, e)$ in Eq. (30) and (32) in practice, we distinguish two cases. We demonstrate them on the example of upper bound.

In the first case, each partially instantiated tuple $c_{1:q}$ that includes node $C_k$, namely $k \leq q$, can be decomposed as $c_{1:q} = z_{1:q} \bigcup c_k'$ so that:

$$P^U(c_k', z_{1:q}, e) = P^U(c_{1:q}, e)$$

The second case concerns the partially instantiated tuples $c_{1:q}$ that do not include node $C_k$, namely $k > q$. In that case, we compute upper bound by decomposing:

$$P^U(c_k', z_{1:q}, e) = P^U(c_k|c_{1:q})P^U(c_{1:q}, e)$$

## Appendix C.  ATB Properties

THEOREM 3.2 *ATB* bounds interval length is upper bounded by a monotonic non-increasing function of $h$:

$$P_{\mathcal{A}}^U(x|e) - P_{\mathcal{A}}^L(x|e) \leq \frac{\sum_{j=h+1}^{M} P(c^j)}{\sum_{i=1}^{h} P(c^i, e) + \sum_{j=h+1}^{M} P(c^j)} \triangleq I_h$$

PROOF. The upper bound on the bounds interval follows from the fact that, $P_{\mathcal{A}}^U(x|e) - P_{\mathcal{A}}^L(x|e) \leq P_{BF}^U(x|e) - P_{BF}^L(x|e)$ and from the definitions of brute force lower and upper bounds given by Eq. (21) and (22). We only need to prove that the upper bound is monotonously non-increasing as a function of $h$.

$$I_{h-1} = \frac{\sum_{j=h}^M P(c^j)}{\sum_{i=1}^{h-1} P(c^i, e) + \sum_{j=h}^M P(c^j)} = \frac{P(c^h) + \sum_{j=h+1}^M P(c^j)}{\sum_{i=1}^{h-1} P(c^i, e) + P(c^h) + \sum_{j=h+1}^M P(c^j)}$$

Since $P(c^h) \geq P(c^h, e)$, then replacing $P(c^h)$ with $P(c^h, e)$ and applying Lemma 3.1, yields:

$$\begin{aligned} I_{h-1} &\geq \frac{P(c^h, e) + \sum_{j=h+1}^M P(c^j)}{\sum_{i=1}^{h-1} P(c^i, e) + P(c^h, e) + \sum_{j=h+1}^M P(c^j)} = \frac{P(c^h, e) + \sum_{j=h+1}^M P(c^j)}{\sum_{i=1}^h P(c^i, e) + \sum_{j=h+1}^M P(c^j)} \\ &\geq \frac{\sum_{j=h+1}^M P(c^j)}{\sum_{i=1}^h P(c^i, e) + \sum_{j=h+1}^M P(c^j)} = I_h \end{aligned}$$

Thus, $I_{h-1} \geq I_h$. $\square$